\documentclass[preprint,12pt]{elsarticle}

\DeclareUnicodeCharacter{2212}{-}
\usepackage{amssymb}
\usepackage{wrapfig,lipsum,booktabs}
\usepackage{amsmath}
\usepackage{xcolor}
\usepackage{graphicx} 
\usepackage{float} 
\usepackage{lipsum} 
\usepackage{booktabs}%
\usepackage{adjustbox}
\usepackage{longtable}
\usepackage{graphicx}
\usepackage{caption}
\usepackage{comment}
\usepackage{algorithm}
\usepackage{algpseudocode}
\usepackage{amsmath}
\usepackage{amsfonts}
\usepackage{amssymb}
\usepackage{array}
\usepackage{textcomp}
\usepackage{stfloats}  
\usepackage{url}
\usepackage{algpseudocode}
\usepackage[colorlinks]{hyperref}
\usepackage[colorinlistoftodos]{todonotes}
\usepackage{verbatim}

\usepackage{url}

\usepackage{fancyhdr} 
\usepackage{hyperref} 

\journal{Nuclear Physics B}

\begin{document}
\begin{frontmatter}

\author{Saja Tawalbeh\corref{cor1}\fnref{label2}}
\ead{Saja.Tawalbeh@uantwerpen.be}

 \cortext[cor1]{Corresponding author}

\title{Towards the Characterization of Representations Learned via Capsule-based Network Architectures}
\author[inst1]{José Oramas}

\affiliation[inst2]{organization={University of Antwerp, Department of Computer Science, sqIRL/IDLab, imec}\\
            ,addressline={Sint-Pietersvliet 7}, 
            city={Antwerp},
            postcode={2000}, 
            country={Belgium}}

\begin{abstract}Capsule Neural Networks (CapsNets) have been re-introduced as a more compact and interpretable alternative to standard deep neural networks. While recent efforts have proved their compression capabilities, to date, their interpretability properties have not been fully assessed.
Here, we conduct a systematic and principled study towards assessing the interpretability of these types of networks. We pay special attention towards analyzing the level to which \textit{part-whole} relationships are encoded within the learned representation.
Our analysis in the MNIST, SVHN, CIFAR{-}10, and CelebA datasets on several capsule-based architectures suggest that the representations encoded in CapsNets might not be as disentangled nor strictly related to \textit{parts-whole} relationships as is commonly stated in the literature.
\end{abstract}


\begin{highlights}
\item This study proposes a principled methodology for assessing the interpretation capabilities of capsule networks.

\item This study aims to verify potential \textit{part-whole} relationships encoded within various components of existing CapsNet architectures.

\item We conduct an empirical analysis to assess the level to which CapsNet-based representations do encode \textit{part-whole} relationships. To the best of our knowledge, this is one of the first efforts conducting this type of study.

\item As part of our methodology, we introduce two methods for the extraction of relevant units in CapsNet architectures i.e., class-agnostic forward path estimation.

\end{highlights}

\end{frontmatter}

\section{Introduction}
\label{inntroduction}
Capsule Neural Networks (CapsNets) \cite{SF17} were recently re-introduced as a more compact and interpretable alternative to deep neural networks i.e., Convolutional Neural Networks (CNNs)
, by modelling spatial hierarchies through \textit{part-whole} relationships. While CNNs rely on pooling operations, which can lose critical positional information, CapsNets utilize groups of neurons that encode both the presence and pose of features, thereby preserving spatial relationships across layers. 
This architecture allows CapsNets to more effectively represent complex transformations and object structures, enhancing interpretability by making the relationships between parts and wholes within an image explicit.
This motivated their introduction in different critical applications including healthcare~\cite{deepika2022improved,afriyie2022classification,WWLLT23}, NLP \cite{kim2020,cheng2022hsan}, object detection~\cite{lin2022feature,yu2021sparse}, salient object detection\cite{liu2022disentangled, liu2024capsule, liu2021part, liu2024deep}, hyperspectral imaging~\cite{Hyperspectral,wang2018hyperspectral,wang2022}, autonomous driving~\cite{HI18,grigorescu2020survey,2020video}, and finance~\cite{sezer2020financial}.
Different from standard convolutional neural networks (CNNs), which arrange neurons in a predefined 2-dimensional manner, a capsule is a group of neurons whose activity vector represents the instantiation parameters of a specific type of entity such as an object or an object part. 
Then, by explicitly providing a mechanism to link specific capsules at neighboring layers, \textit{part-whole} relationships are modelled. Based on these mechanisms ensembles of capsules are capable of modelling higher level semantic concepts, e.g. objects and scenes, with a proper spatial arrangement.
Recent efforts~\cite{mukhometzianov2018capsnet,JL20} have shown that CapsNets are capable of obtaining comparable results with respect to their convolutional counterparts while requiring less 
parameters. Thus, confirming their compression capabilities. However, while it is theoretically expected that the activities of the neurons within an active capsule represent relevant properties of the data (e.g. deformation, velocity, hue, texture, etc.), to date, this built-in interpretable characteristic has not been systematically assessed.

Efforts towards improving the intelligibility of systems based on deep models have been oriented in two fronts. Either, by providing insights into what a model has actually learned (model interpretation), or by providing justifications for the predictions made by these models (model explanation). In the last decade, a significant amount of work~\cite{FV17,GR16,HA16,ZF14} has been conducted towards explanation. In contrast, the amount of efforts~\cite{BK17,OW19} around the interpretation task is much more reduced. A more critical trend has been observed in the context of CapsNets, where efforts toward their interpretation are almost non-existent. Putting the previous points together paints a worrying picture. CapsNets are a type of model that is proving effective and is gaining attention in the context of several critical applications. However,  when compared with its CNN-based counterparts, this type of model is not well studied and their inner-working are not necessarily well understood. 

Starting from these observations, we aim to experimentally assess the interpretability capabilities of CapsNets. More specifically, we verify whether the internal representations of CapsNets do encode features that are both relevant for the data they were trained on, and critical for the performance on the task at hand, e.g. classification. This paper puts forward the following contributions:
i) We propose a principled methodology for assessing the interpretation capabilities of capsule networks.
This aims at verifying the \textit{part-whole} relationships by looking at the inner workings of the CapsNet across all its layers.
ii) We conduct an empirical analysis to assess the level to which CapsNet-based representations do encode \textit{part-whole} relationships. 
To the best of our knowledge, this is one of the first efforts conducting this type of study.
iii) As part of our methodology, we propose two methods for the extraction of relevant units in CapsNets. These units lead to comparable performance.

\begin{figure*}[t!]
   \centering
\includegraphics[width=\linewidth]{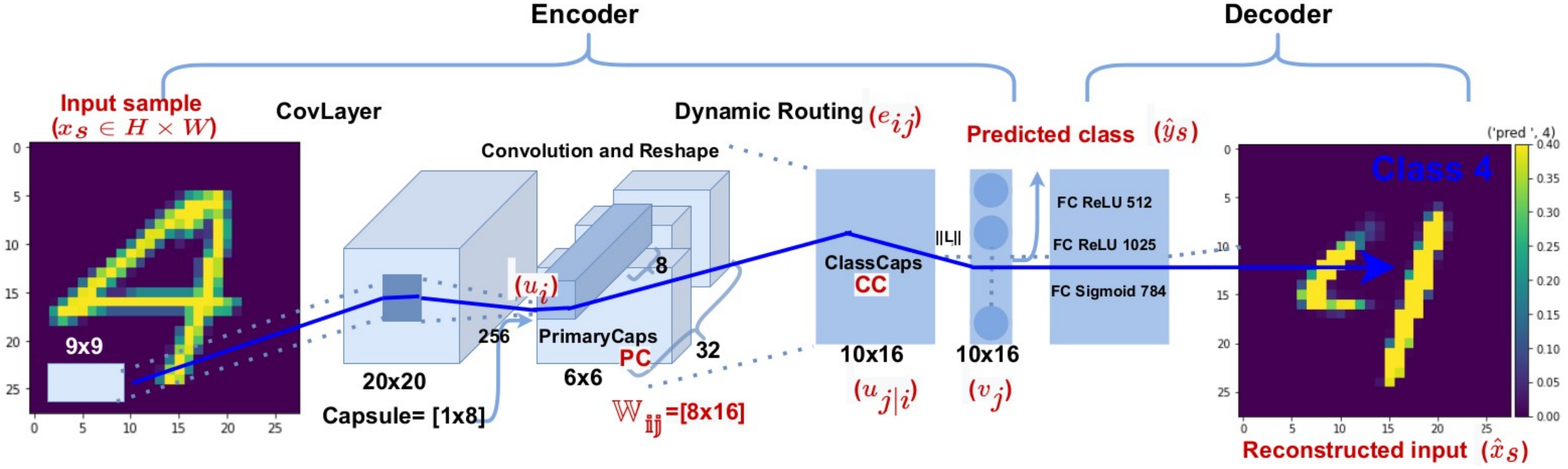}
  \vspace*{-0.12in}
   \caption{Capsule network architecture from~\cite{SF17}. The network consists of 3 layers; \textit{Conv} layer, \textit{PC} layer, and \textit{CC} layer and a decoder.
  \label{fig:capsuleArchitecture}}
\end{figure*}
\section{Related Work}
\label{sec:relatedWork}

Understanding how deep complex models operate internally have received significant attention in recent years~\cite{ZF14,SV14,GR16,selvaraju2017grad}. 
Most of the methods in the literature operate in a {post-hoc} manner, i.e. they provide interpretation/explanation capabilities on a pre-trained model. More recently a new trend has emerged which focuses on the design of methods or learning algorithms that aim to produce models that are interpretable. Thus producing models that are interpretable-by-design~\cite{ZW18}.

Related to this trend, \cite{SF17} proposed CapsNets. A characteristic aspect of this model is that the learned representations encode \textit{part-whole} relationships from features present in the data.
\cite{LQ19} analyzed their proposed CapsNet-based recommendation model (CARP) which verifies whether it was able to detect suitable reasons and their effects. This was done by retrieving the top-$k$ phrases which are used to interpret rating behavior. 
In a similar manner, \cite{WH20} proposed a multi-head attention layer with capsules that capture the semantic aspects which resemble the global interpretation with respect to the entire dataset. Moreover, they explain the output of the network for a given input by referring to the primary capsule that strongly agrees with the class capsule.

In the medical context, \cite{SG18} proposed a modified CapsNet model for automatic thoracic disease detection. 
Grad-CAM~\cite{selvaraju2017grad} visualizations were produced to inspect regions of interest that are considered critical for assessing the location of the disease. The visualizations were produced from the last layer. In contrast, our methodology produces visualizations from multiple layers to analyze the information each unit in these layers encodes i.e., \textit{part-whole}.
In the domain of hyperspectral imaging, \cite{SH21} proposed a two stage model for vegetation recognition. The interpretability of their model was improved by using a capsule-based stage for learning a hierarchical representation from inputs. This capsule-based stage was composed of enriched low-level feature representations computed during the first stage. The interpretation capabilities of this model were assessed by measuring the correlation between intermediate features in the model and annotations in auxiliary datasets.
\cite{JC18} proposed a CapsNet-based method for protein classification and prediction. Outputs produced by this model are explained by following an input-modification method in which information about some "atoms" of a given input is modified and its effect on the output is measured. \cite{SM18} analyzed the behavior of CapsNets ( by \cite{SF17}) to identify the relevant activation path defined by the bi-product of the routing procedure and use it as an explanation for the network.
Similar to \cite{SF17,SM18, JC18}, we produce reconstructions of the input images by perturbing the output vector $v_j$. In contrast, we produce extensive visual reconstructions by systematically altering $v_j$ based on a defined interval estimated in a principled manner.

\cite{JL20} proposed the interpretable iCaps model which produces explanations of classification predictions based on relevant information in active capsules. This is achieved by an additional supervised approach for representation disentanglement and a regularizer that ensures low redundancy on the concepts encoded in the model.
Differently, our methodology examines the disentanglement within the original CapsNet architectures rather than proposing a new architecture designed to enforce disentanglement, as explored in \cite{JL20}.
Previously \cite{BA20,SM18} introduced interpretation methods to verify the path based on the dynamic routing procedure introduced in~\cite{SF17}. 
More recently, new efforts emerged, questioning the hierarchical relationships encoded in CapsNets. Along this line, \cite{MKGS23} argues that CapsNets do not exhibit any theoretical properties suggesting the emerging of parse trees in their encoded representation. They discussed how a parse tree structure is crucial when capsules are expected to serve as nodes and their connections function as edges. 

Inspired by \cite{BA20,SM18}, we follow a similar approach in our experiments to obtain and interpret the relevant connections between layers. While previous efforts primarily focused on the routing procedure alone, our approach extends this by examining the impact of both the routing procedure and capsule activations across different stages of the network. This broader perspective allows for a more comprehensive understanding of information flow within the network and reveals how these interactions collectively contribute to the final prediction.
In line with \cite{WH20} we consider local and global interpretation from models trained on visual data. Different from it, we analyze the internal behavior, including the convolution layer, which is crucial for the feature extraction stage. This layer-wise exploration allows us to understand how features unfold through different stages of the network. 
Moreover, unlike prior work, we analyze forward and backward path estimation alternatives. This dual-path investigation enables a comprehensive evaluation of how information is propagated through the network during inference (forward pass) and how gradients flow during training (backward pass).
Similar to \cite{SG18} we also explain model predictions, visually, by producing heatmaps. In a different manner, we produce heatmaps for the top relevant units located at the intermediate layer.
Different from iCaps \cite{JL20} and all previous efforts, we analyze the internal behavior of the CapsNet architectures. We analyzed the drop in performance when relevant filters and capsules are removed in such a network. 
Similar to \cite{MKGS23}, we introduce a systematic study towards assessing the interpretability of several Capsule Architectures. However, we propose a more general protocol that extends beyond examining the connections (routing connections) between various levels of capsule layers.
Our protocol consists of perturbation analysis and extracts the relevant features that define the relevant paths connecting the inputs and outputs. We also assess the emergence of the \textit{part-whole} relationships by generating visual heatmaps. This is a key difference in our methodology compared to others. However, we observe that related efforts typically focus on only one or two aspects in their methodologies.

\cite{3DZYGG23} proposes a method in the domain of 3D object representations to analyze CapsNet autoencoders inspired  by~\cite{KS19}. This method aims to construct a structured latent space and posits that a specific component associated with an object can be separated into independent sub-spaces.
Similar to our method, \cite{3DZYGG23}, this method scrutinizes the behavior of CapsNet by demonstrating its capacity to reconstruct ground-truth parts w.r.t their spatial locations.

Several efforts have been proposed to enhance salient object detection through an advanced capsule routing algorithm, each addressing the challenges of object wholeness and complexity in unique ways. 
For instance, \cite{liu2024deep} introduces a multi-stream capsule routing method that prioritizes important features via a belief score for each stream, effectively tackling the issue of poor object wholeness commonly observed in existing methods. This aligns with the works of \cite{KS19,3DZYGG23}, which also explore hierarchical relationships within salient regions using unsupervised learning. 
\cite{liu2021part} adopts a two-stream strategy aimed at reducing network complexity and redundancy during capsule routing. By investigating part-object relationships, they generate a capsule wholeness map that facilitates the integration of multi-level features to build a more cohesive saliency map.
In another notable contribution, \cite{liu2024capsule} presents a residual routing algorithm designed to enhance the routing between capsule layers, which reduces computational complexity. This method produces saliency predictions and 3D reconstructions to explore part-whole relationships. 
Additionally, the method proposed in \cite{liu2022disentangled} examines disentangled representation capsule routing on salient object detection. It also accelerates part-object relational saliency.
The efforts mentioned above differ significantly from our methodology, as they apply CapsNets to the task of salient object detection, focusing on identifying the most noticeable segment or part within an image. This contrasts with our objective, which is to verify whether \textit{part-whole} relationships exist within the units that compose the CapsNet architecture.

\begin{figure}[t!]
\centering
\includegraphics[width=\linewidth]{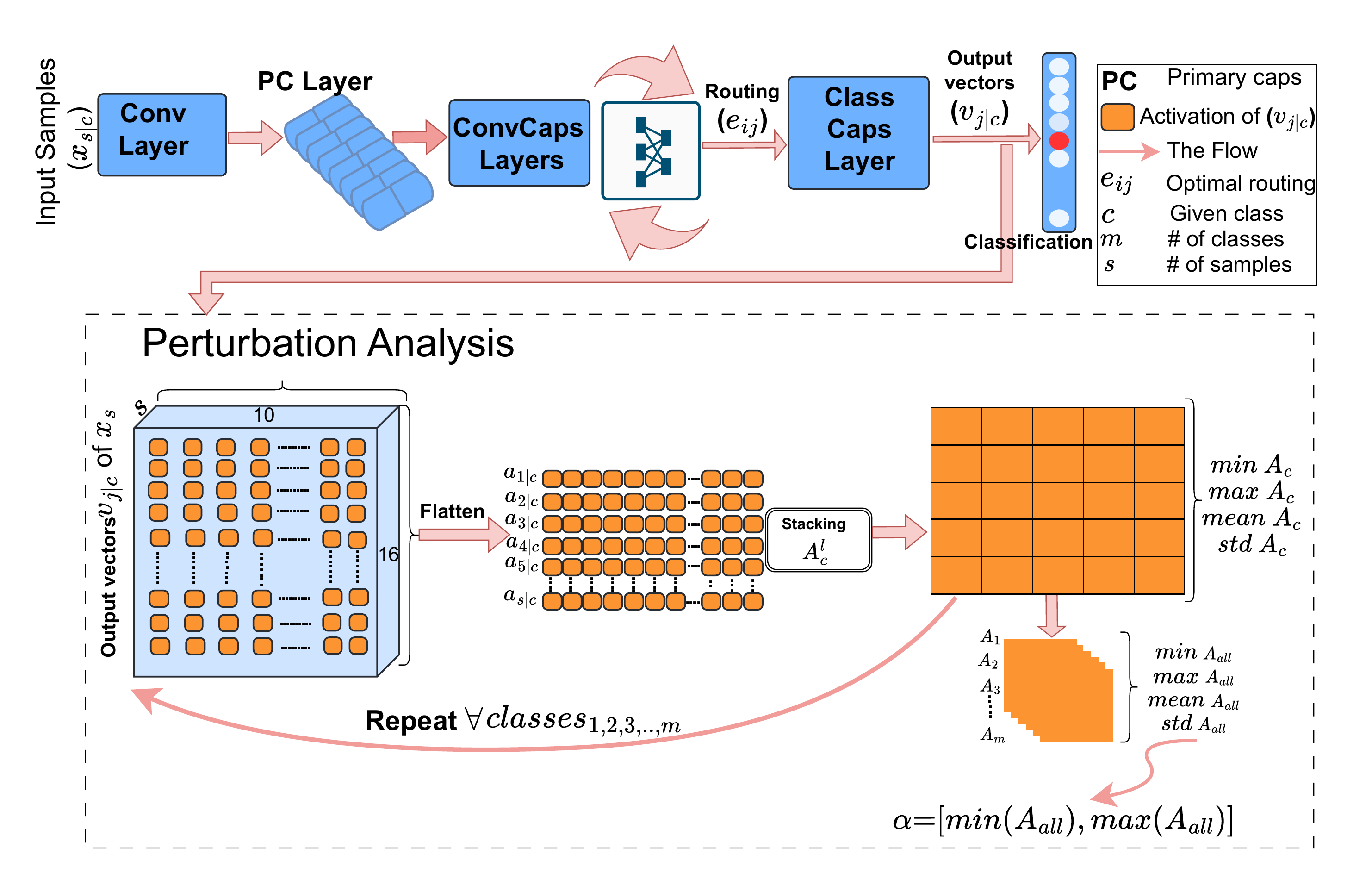}
\vspace*{-0.12in}
\caption{An overview of the computation of the interval $\alpha$, applied to each layer. This interval is utilized in our perturbation analysis (Section 4.1).\label{fig:perturbationAnalysisDigram}}
\end{figure}
\begin{algorithm}[b!]
\caption{Perturbation Analysis}\label{alg:Perturb}
\begin{algorithmic}[1]
    \State \textbf{Input:} $x_s$ (a sample of a given class)
    \State \textbf{Output:} $\hat{v}_j$ (perturbed vector)
    \State \textbf{Definitions:} $a_{s,c}$ (activations), $v_j$ (output), $d$ (dimensions), $f$ (features), $\alpha$ (perturbation ranges)
    \For{each $x_s$ in dataset}
        \For{each filter}
            \State extract features $f$
        \EndFor
        \State $a_{s,c} \gets$ flatten and concatenate filters
        \State compute min, max, mean, std of $a_{s,c}$ per class
        \State $A_{c} \gets$ first-order statistics $\eta^l_c$ (min, max, mean, std)
        \State $A_{all} \gets$ global statistics over all classes
    \EndFor
    \State $\alpha \gets [\text{min}(A_{all}), \text{max}(A_{all})]$
    
    \While{$d < 16$}
        \State $\hat{v}_j \gets$ copy($v_j$)
        \For{each $r$ in $\alpha$}
            \State $\hat{v}_j[d] \gets r$
        \EndFor
        \State $d \gets d + 1$
    \EndWhile
\end{algorithmic}
\end{algorithm}

\section{Overview: Capsule Networks}
\label{sec:OverviewCapsuleArchitecture}
Fig.~\ref{fig:capsuleArchitecture} presents 
an overview of the CapsNet architecture proposed by \cite{SF17}. In a CapsNet architecture, each capsule is defined by a set of neurons (represented by the activity vectors denoted by  $u_i \in \mathbb{R}^{
 d^{l} {\times} d^{l+1}}$, where $d^l$ and $d^{l+l}$ are the dimensions of the primary and class capsule layers respectively) along with instantiation parameters $T$. 
 These parameters represent the features of the entities in the desired dataset such as pose (position, size, orientation), lighting, deformation, and velocity. The length of the output vector $v_j$ indicates the likelihood of occurrence of an entity. A standard CapsNet architecture (as proposed in \cite{SF17}) consists of two parts (Fig.~\ref{fig:capsuleArchitecture}). First; an \textit{encoder} which is composed of three layers; a convolutional layer (\textit{Conv}), a primary capsule layer (\textit{PC} ), and a class capsule layer (\textit{CC}). Second, a \textit{decoder} which aims a reconstructing the input.

\begin{figure*}[t!]
\centering
\includegraphics[width=\linewidth]{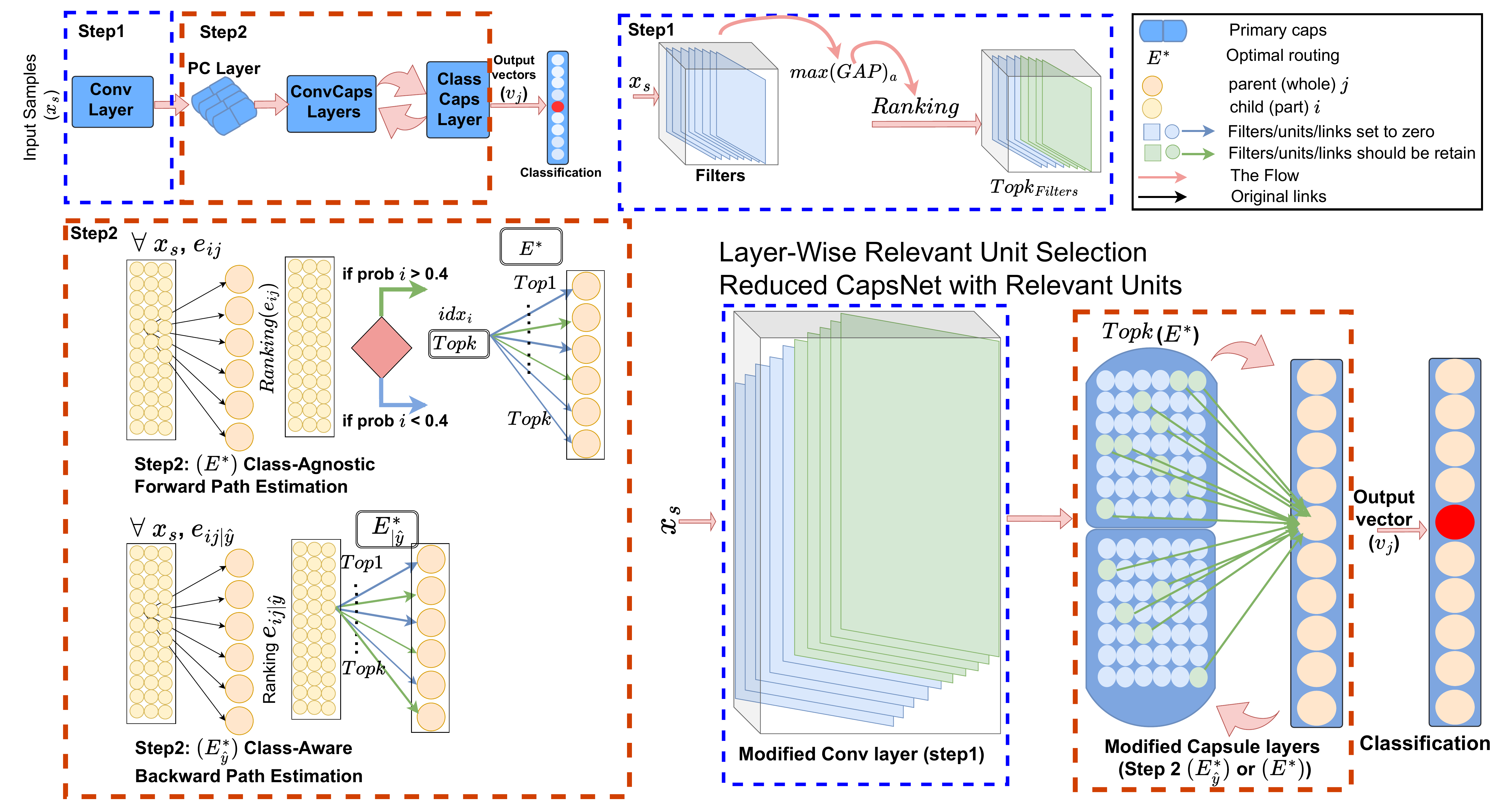}
\vspace*{-0.10in}
\caption{Overview of the proposed path estimation methodology. The input is passed through a trained CapsNet to obtain activations from different layers. In the top left, the original CapsNet is illustrated. The top right depicts the process of selecting relevant units within the \textit{Conv} layer (blue rectangle). The bottom left outlines the selection of relevant units between the \textit{PC} and \textit{CC} layers (red rectangle). Finally, the bottom right shows the reduced CapsNet, through which each input is processed for further analysis (Section 4.2). \label{fig:pathEstimationDiagram}}
\end{figure*}
The \textit{encoder} part starts with a standard \textit{Conv} layer which aims to obtain the initial features (e.g., edges and lines) from an input image $x_s \in \mathbb{R}^{H {\times} W}$. The \textit{Conv} layer is followed by a \textit{ReLU} activation function.
The \textit{PC} layer is represented by 32 \textit{PC} capsules which aim to detect high-level features (e.g., part of an object) based on features extracted by the \textit{Conv} layer. \textit{PCs} perform 8 convolution operations each time. Each $PC_i$ in the \textit{PC} layer (\(i \in [1 \leq i \leq N]\), with $N$ the number of \textit{PC} capsules) encodes $T$ by vector $u_i$. Consequently, $CC^j$ in the \textit{CC} layer (\(j \in [1 {\leq} j {\leq} M]\), with $M$ being the number of discrete classes to be predicted (e.g., $M{=}10$ in case of the MNIST digits). The \textit{CC} layer receives an input from the \textit{PC} layer whose size depends on the dataset being used (e.g., $6{\times}6{\times}8{\times}32$ in case of the MNIST digits).
Then, an affine transformation is applied via the transformation matrix $\mathbb{W}_{ij} \in\mathbb{R}^{d^{l} {\times} d^{l+1} }$. This type of transformation is a unique procedure to CapsNet since it considers both the missing spatial relationships and the relationships between parts in relation to the object.
A product between the transformation matrix $W_{ij}$ and capsules in the $PC$ layer is computed. 
This transforms $u_i$ to a vote/predicted vector $\hat{u_{j|i}} \in\mathbb{R}^{d^{l+1}}$ towards the $j^{th}$ capsule.  
The predicted vector $\hat{u_{j|i}}$ indicates how the $i^{th}$ capsule from the \textit{PC} layer contributes to the $j^{th}$ capsule in the \textit{CC} layer. 
Coupling coefficients $e_{ij} \in\mathbb{R}^{d^{l} {\times} p^{d+1}}$ are added, and further computed, to indicate the forward path/flow (links) from $PC_i$ capsule at $l$ to the $CC_j$ capsule at $l+1$.
$e_{ij} \in[0,1]$;  
$i \in \{1, 2, ..., n\}$.
In the literature, this procedure is usually referred to as \textit{dynamic routing} \cite{SF17}. 
In CapsNets, \textit{CC} is the classification layer, where each capsule vector $v_j$ indicates a single class.
The predicted class $\hat{y_s}$ is obtained by looking at the highest predicted probability $v_j \in\mathbb{R}^{d^{l+1}}$ (see Fig.~\ref{fig:capsuleArchitecture}).
The \textit{decoder} consists of several fully connected layers. It aims to reconstruct the input example $x_s$ from the vector $v_j$ that is produced by the CC layer. The decoder output is the reconstruction $\hat{x}_s$.
\section{Methodology}
\label{sec:methodology}
This section describes the methodology for assessing the interpretability properties of the representation encoded in CapsNets. The methodology consists of two parts.
First, a hollistic perturbation analysis in which various parts of the architecture are ablated to assess their impact on classification performance.
Second, during the forward pass, the activation paths linking input and output are analyzed with the goal of verifying the most relevant features in each layer and their influence on modelling \textit{part-whole} relationships.
\subsection{Perturbation Analysis}
\label{sec:featureAnalysisRelevantPerturbation}
Given a trained  CapsNet model $F$ that takes an input $x$ and produces a class label $\hat{y}$ as output, i.e., a classifier. The first step is to push every example $x_{s,c}$ belonging to class $c$ and extract the activations $a^l_{s,c} \in \mathbb{R}^{w {\times} h {\times} d}$ for every internal layer $l$. 
Then, after computing the activations $a^l_{s,c}$ of every example $s$ in the dataset, flattening them and concatenating them on top of each other, a matrix 
$A^l_c = [ a_{1,c}^l ; a_{2,c}^l ; ... ; a_{s,c}^l ]$ is defined with $A^l_{c} \in\mathbb{R}^{[S\prime {\times} A\prime]}$ where $S\prime$ refers to number of training examples corresponding to a class $c$ and $A\prime$ refers to total number of ($1$D) flattened activations.

With $A^l_c$ in place, first-order statistics $\eta^l_c = [min(A^l_c) ; max(A^l_c) ; mean(A^l_c);$ $std(A^l_c)]$ $ 
\in\mathbb{R}^{[4 {\times} A\prime]}$ 
are computed in a column-wise manner. 
To complement this, a similar $\eta$ matrix is composed of the first-order statistics across the whole dataset $A^l_{all} =[ A^l_1 ; A^l_2 ; ... ; A^l_M ] \in\mathbb{R}^{[D {\times} A\prime]}$, where $D$ refers to the number of examples in the dataset.
%
\begin{algorithm}[t!]
\caption{Class-Agnostic Forward Path Estimation}\label{alg:relevantPCFP}
\begin{algorithmic}[1]
    \State \textbf{Input:} $x_s$ (a sample of a given class)
    \State \textbf{Definitions:} $k$ (top-$k$ relevant units), $e_{ij}$ (coupling coefficients), $E*$ (optimal coupling coefficients)
    \While{$x_s$ in dataset}
        \State extract $e_{ij}$ for all classes
        \State rank $e_{ij}$ with corresponding capsules
        \If{$e_{ij} > 0.4$}
            \State \textit{E*} $\gets$ index[$e_{ij}$]
            \State $k \gets$ select(top-$k$(\textit{E*})) 
        \EndIf
    \EndWhile
\end{algorithmic}
\end{algorithm}
From here, we define the interval $\alpha = [ min(A_{all}^l), max(A_{all}^l) ] $ which represents the empirical range of the activation space of a given unit at layer $l$ in the CapsNet.
%
\begin{table}[b!]
  \centering
    \caption{Mean classification accuracy on the MNIST, SVHN, CIFAR{-}10, and CelebA datasets for the original dense CapsNet architectures proposed by \cite{SF17,HSF18} and a sparser version based on the identified (\textit{Forward} or \textit{Backward}) activation paths (which only propagate the relevant units) \label{tab:performanceDataset}}
  \begin{adjustbox}{width=\textwidth}
  \begin{tabular}{cl|llllllll}
    \toprule
    \multicolumn{2}{c}{Capsule Type}
     & \multicolumn{2}{c}{MNIST}  &  \multicolumn{2}{c}{SVHN} &  \multicolumn{2}{c}{CelebA}&  \multicolumn{2}{c}{CIFAR{-}10}\\ 
    \cmidrule(r){1-10}
     &   & Valid  & Test & Valid & Test & Valid & Test& Valid & Test \\
    \midrule
      & Dense (original) & 99.9  & 99.1 & 96.7  & 91.0 & 93.0 & 92.0 & 93.6 & 84.3  \\
     
    \textbf{Dynamic Routing} & Backward Path & 76.9  & 78.5 &  89.3 & 87.2 & 81.4 & 80.6 & 99.9 & 99.9  \\
    
    & Forward Path & 95.2 & 95.1 & 96.0 & 88.3 & 89.1 & 86.2 & 99.9 & 99.7 \\
    \midrule
    & Dense (original) & 98.1 & 98.0 &81.4 & 80.0 & 85.7 & 85.7 & 73.4 & 72.2 \\
    \textbf{Em Routing} & Backward Path & 98.6 & 97.6 & 84.3 & 78.7 & 85.6 &85.1& 72.3& 71.2 \\ 
    &Forward Path & 98.8 & 97.8 & 83.3 & 78.9 & 85.7 & 85.2 & 72.9 & 65.0 \\
    \bottomrule
\end{tabular}
  \end{adjustbox}
\end{table}
By estimating the range $\alpha$ in this way, we ensure the proper analysis of the activation space of different units located at different layers of the model.
This differs significantly from previous efforts~\cite{SF17,amer2020path,SM18}, which use heuristically-defined  $\alpha$ values. 
We apply a perturbation analysis on $v_j$ (output of the \textit{CC} layer) based on $\alpha$ for a given class $c$. We systematically replace each dimension of $v_j$ by an uniformly-sampled value $\xi$ in the range $\alpha$, which produces a new vector $v'_{j}  \in\mathbb{R}^{ d^{l+1}}$.
Then, $v'_{j}$ is fed to the decoder to produce a reconstruction $\hat{x}_{s,c}$.
In our analysis, the reconstructed input $\hat{x}_{s,c}$ has two purposes. It is used to conduct a systematic qualitative assessment on how variations in one of the dimensions of $v_j$ lead to different reconstructions. In addition, the effects of these variations can be quantitatively assessed by measuring the difference in classification performance that is obtained when $\hat{x}_{s,c}$ are fed to the CapsNets as inputs. 
A detailed description is provided in Algorithm~\ref{alg:Perturb}. Additionally, Fig.~\ref{fig:perturbationAnalysisDigram} illustrates the flow underlying the developed perturbation analysis.
\subsection{Layer-Wise Relevant Unit Selection} 
\label{sec:layerwiseOcclusion}
This method is aimed at detecting the relevant features/units that define activation paths in a given network.
This is achieved by probing one layer at a time and verifying how the selection/suppression of internal units in such layers affects classification performance.
Fig.~\ref{fig:pathEstimationDiagram} provides a detailed overview of the defined activation paths methodology within the network, specifically illustrating the procedure for selecting relevant filters in the \textit{Conv} layer and verifying the units in the Caps layers.

For \textit{Conv} layers, global average pooling (GAP)~\cite{LinCY13ICLR2014} is computed on the output channel produced by each filter. The filters with the highest GAP values, i.e. with the highest average activation, are assumed as the most relevant in the layer. 
Given this relevance rank, as defined by the GAP score of each filter, we select the top-$k$ most relevant filters via cross-validation. 

\begin{algorithm}[t!]
\caption{Class-Aware Backward Path Estimation}\label{alg:relevantPCBP}
\begin{algorithmic}[1]
    \State $x_s \gets$ (a sample of a given class)
    \State $k \gets$ top-$k$ relevant units
    \State $\hat{y} \gets$ predicted label
    \State \textit{E*} $\gets$ optimal coupling coefficients
    \While{$x_m$ \text{in} \textit{dataset}}
        \State $\hat{y} \gets$ CapsNet($x_s$)
        \State $e_{ij} \gets$ extract(coefficients of the predicted class $\hat{y}$) 
        \State $e_{ij} \gets$ ranked($e_{ij}$)
        \State \textit{E*} $\gets$ select(top-$k$($e_{ij}$))
    \EndWhile
\end{algorithmic}
\end{algorithm}

%
More precisely, given trained CapsNet and a validation set, we feed every example and gradually increase the number ($k$) of considered filters that are considered. During inference, the output/response of the selected filters is preserved while that of the rest is suppressed, i.e. set to zero.
After repeating this procedure for several $k$ values, the lowest $k$ value which led to the highest classification accuracy is adopted.
We consider these $k$ relevant filters, in the \textit{Conv} layer, as the starting point of the activation path.
For capsule-based layers, we apply a slightly different procedure. As mentioned earlier, CapsNets are characterized by the routing algorithms~\cite{mukhometzianov2018capsnet} which determine how activations flow internally across capsule layers.
Taking advantage of this, the "joint" capsules linking the \textit{PC} and \textit{CC} layers are identified. This is done based on the coupling coefficients $e_{ij}$ that determine the routes across capsule layers. 
Connecting the most relevant filters in the \textit{Conv} layer with the "joint" capsules linking \textit{PC} and \textit{CC} produces a complete path between the input $x_{s}$ and the prediction $\hat{y_s}$. 
It is worth mentioning that during inference, some capsules from the $PC$ layer that hold irrelevant features (for a given class)  may appear to be active \cite{BA20}. However, following the routing algorithms, these capsules are eventually routed to inactive capsules in the $CC$ layer. This prevents them from having a significant effect on the prediction $\hat{y_s}$.
Similar as within the \textit{Conv} layer, the number $k$ of selected units is estimated via cross-validation.

On the basis of the procedure described above, we propose two methods to define activation paths on CapsNets. These methods are characterized by the direction, i.e. \textit{forward} or \textit{backward}, that is followed to estimate the path. These directions are closely related to the awareness of the method on the class $\hat{y_s}$ predicted by the model.\\
\begin{figure}[t!]
\centering
\vspace*{-0.10in}
\includegraphics[width=\linewidth]{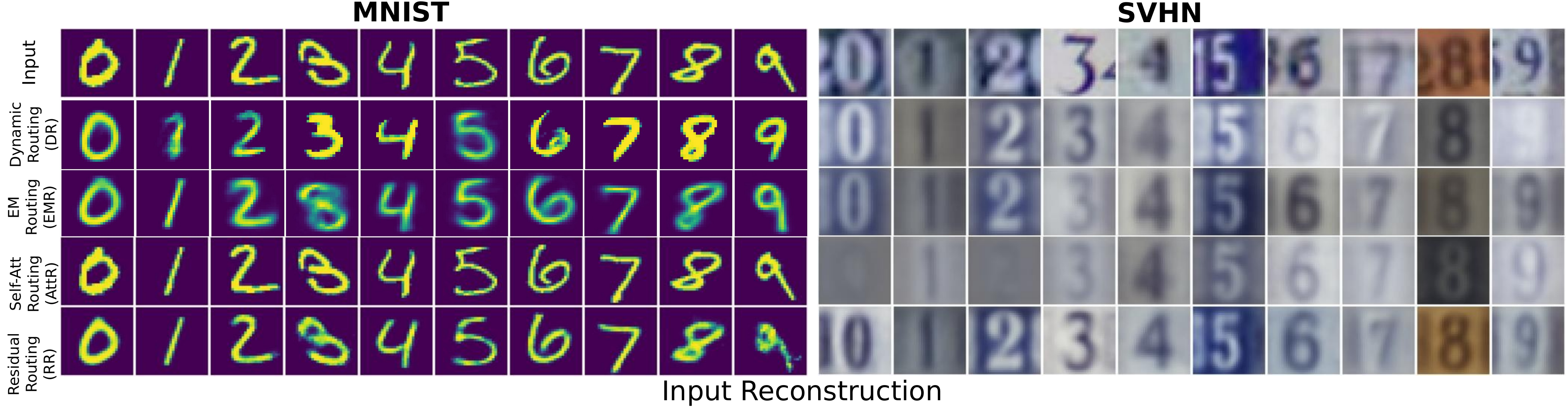}
\vspace*{-0.14in}
\caption{Class reconstructions with different architectures on MNIST and SVHN datasets. For
comparison, we present the input sample $x_s$ data in the first row and the reconstruction results (prediction) in the rest of the rows. Even with different architectures and training settings, all models could embed different properties of the input digits keeping only important details from both datasets.\label{fig:perWRTRelatedMethods}}
\vspace*{-0.10in}
\end{figure}
\textbf{Class-Agnostic Forward Path Estimation. }During the forward pass, the routing procedure defines the path of the predicted vector $\hat{u}$ which represents the relationship between a child (part) $i$ and a possible parent (\textit{whole}) $j$ capsule where we ignored the target classes.
This is achieved by following the internal sequence of units (capsules) that maximize the internal activation flowing forward.
In particular, we solve Eq.~\ref{eq:forwardPass} by passing $x_s$ through the CapNet to obtain the corresponding coefficients \textit{$E^*$}. 
By solving Eq.~\ref{eq:forwardPass}, the optimal \textit{$E^*$} indicates the agreement to the magnitude of $v_j$ is maximized.
%
\begin{equation}
\vspace*{-0.08in}
E^* = \sum_i^N(max(F(\hat{u}))) = \sum_i^N(max(\sum_i^N e_{iM} \hat{u}_{M|i})) \label{eq:forwardPass}
\vspace*{-0.04in}
\end{equation}
In other words, this optimization leads to identifying the relevant capsules that should be activated in the $PC$ layer. Then, we find the relevant capsules of an example $x_s$ based on the coupling coefficients $e_{ij}$ with the highest values. The term $e_{iM} \hat{u}_{M|i}$ refers to coefficients that agree on what could be the predicted class for the input $x_s$.
These steps are described in detail by following the procedure outlined in Algorithm.~\ref{alg:relevantPCFP} for CapsNet architectures. As explained earlier, this process is designed to determine the optimal routing paths (\textit{$E^*$}).\\
\begin{figure*}[t!]
   \centering
    \includegraphics[width=0.92\linewidth]{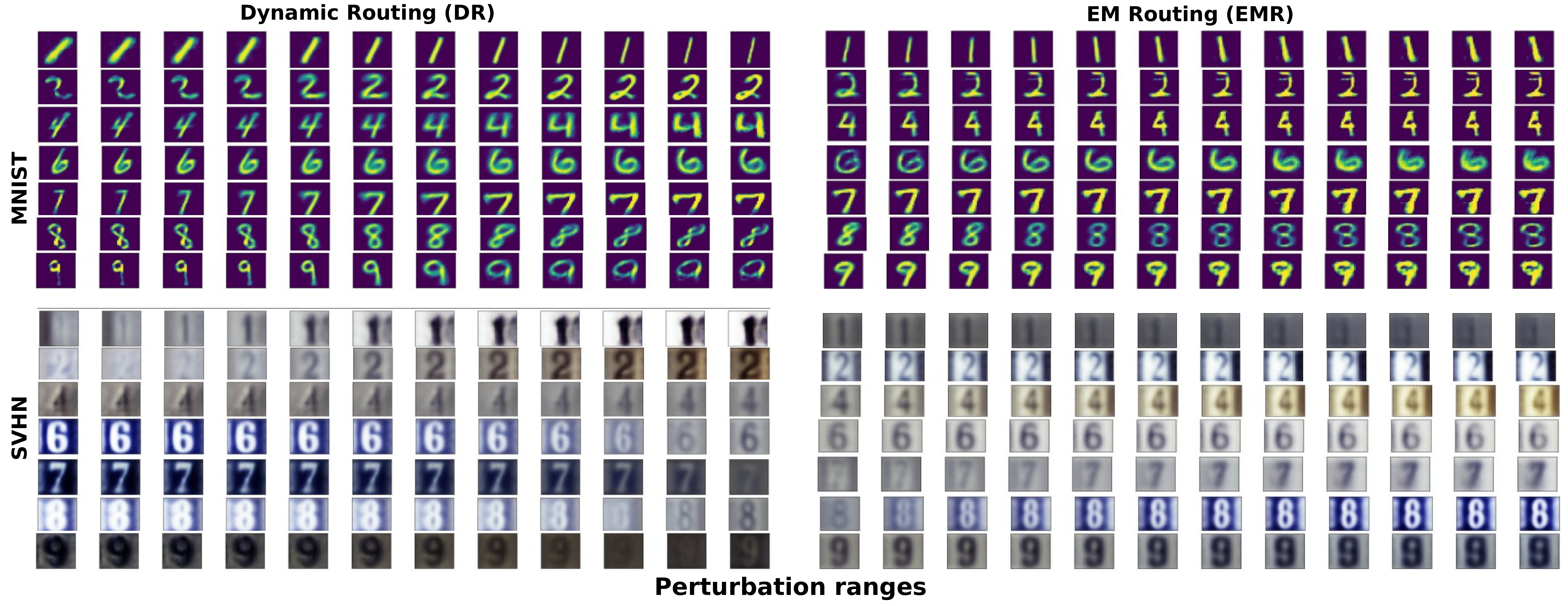}
    \vspace*{-0.11in}
    \caption{
    Qualitative examples of reconstructed inputs as the vector $v_j$ is perturbed based on the overarching ranges (the interval $\alpha$) across classes. It is noticeable in some cases, multiple visual features are modified at the same time via the perturbation of a single dimension of the vector of the active class capsule on both CapsNet architectures.}
    \label{fig:pertubation_Expr}
    \vspace*{-0.10in}
\end{figure*}
\textbf{Class-Aware Backward Path Estimation. }In contrast to the previous method, which defines an activation path in a class-agnostic fashion, here the coefficients \textit{$E^*$} are obtained based on the predicted class $\hat{y}_s$. 
%
\begin{equation}
\vspace*{-0.05in}
E^* = max(f(\hat{u})_{|\hat{y}_s}) = \sum_i^N(max(\sum_i^N e_{i\hat{y}} \hat{u}_{\hat{y}|i})_{|\hat{y}_s})
\label{eq:backwardPass}
\vspace*{-0.05in}
\end{equation}
\begin{figure}[b!]
   \centering
  \includegraphics[width=0.92\linewidth]{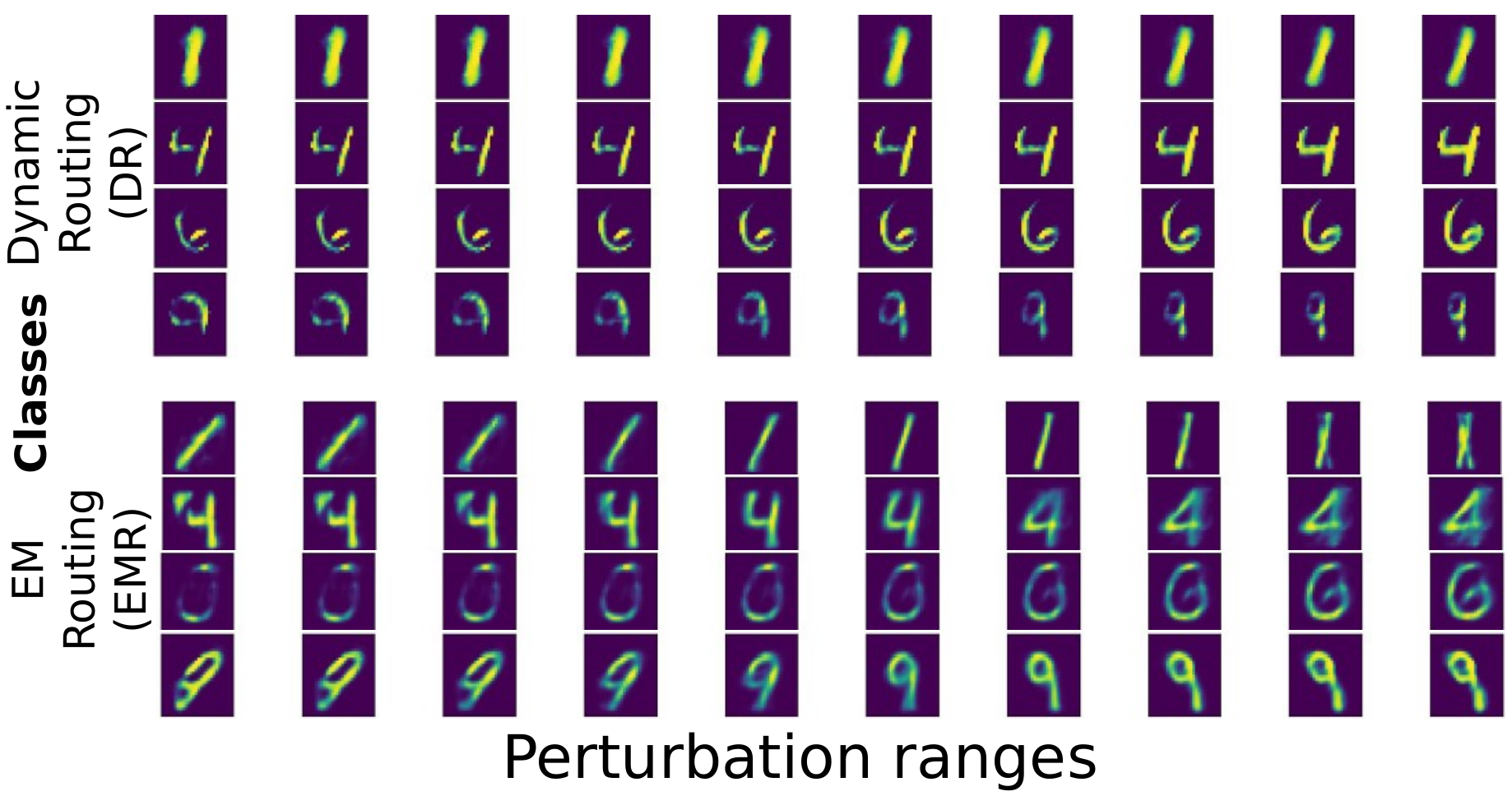}
  \vspace*{-0.14in}
   \caption{Class reconstruction with different perturbations on vector $v_j$ (following the heuristic interval) of the active class capsule obtained from the original CapsNets architectures. The reconstruction images are rotated in different angles,
along with a slight deformation. The perturbation parameter is tweaked from $[−0.25, 0.25]$ by intervals of $0.05$ which is defined by \cite{SF17}.
  \label{fig:pertubationExprRW2}}
\end{figure}
\begin{figure*}[t!]
\centering
\vspace*{-0.10in}
\includegraphics[width=0.90\linewidth]{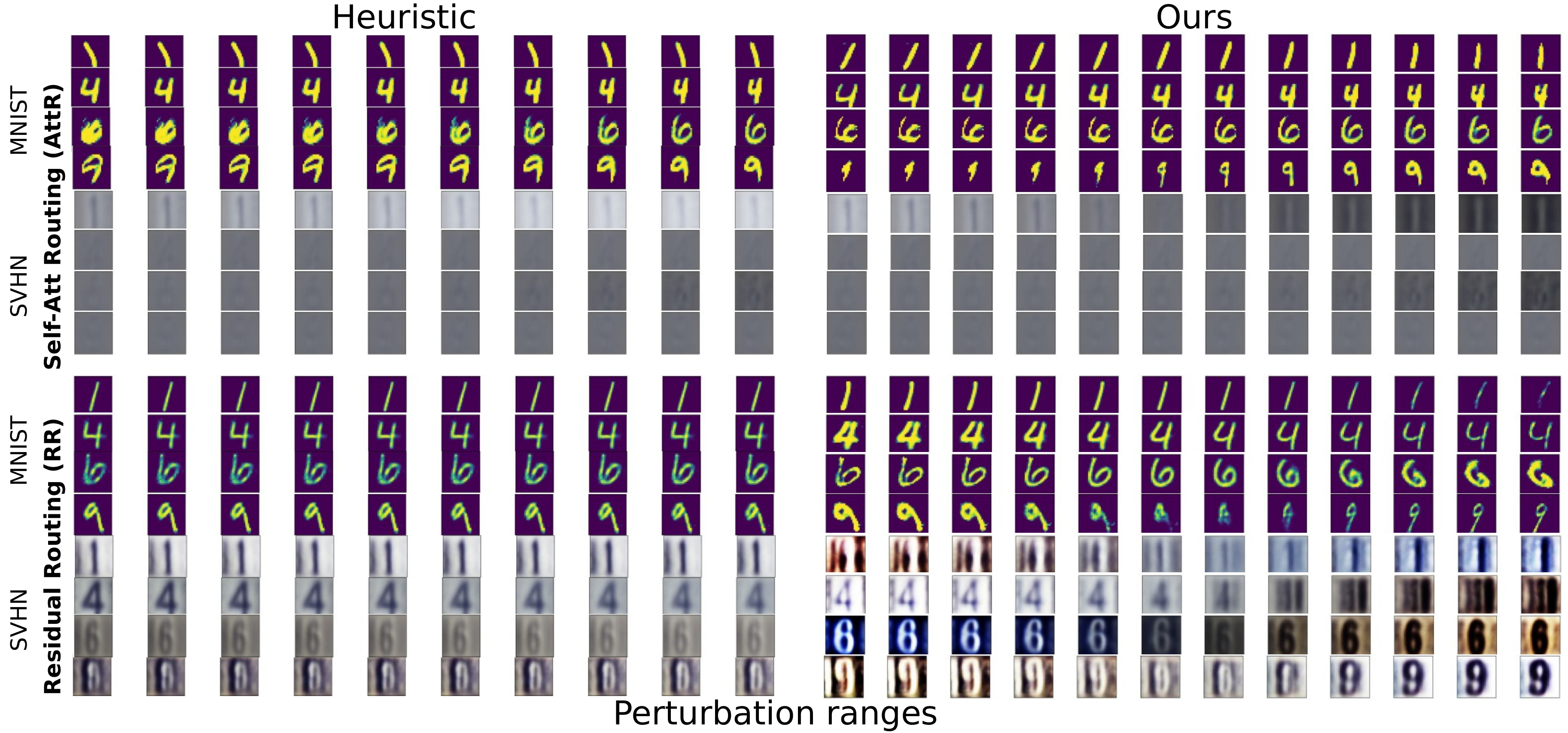}
\vspace*{-0.12in}
\caption{Class reconstruction with different perturbations on vector $v_j$ of the active class capsule obtained from recent CapsNets architectures \cite{MSC21,liu2024capsule}. The reconstruction images are rotated in different angles,
along with a slight deformation. The perturbation parameter is tweaked from $[−0.25, 0.25]$ by intervals of $0.05$ which is defined by \cite{SF17}. We show results for perturbations using the heuristic interval (left) and our proposed interval based on Section 4.1 (right).\label{fig:pertubationExprRWV1}}
\end{figure*}
More specifically, each input example $x_s$ is fed to the CapsNet to produce a predicted class $\hat{y}_s$. Then, in a reverse manner, we extract the $e_{ij}$ of the prediction of class  $\hat{y}_s$. 
Another difference with the previous method, which included all class coefficients, is that this method ignores the $e_{ij}$ that might be relevant to other classes. 

In practice, this is a achieved by solving Eq.~\ref{eq:backwardPass} given \textit{$\hat{y_s}$} where the term $(e_{i\hat{y}}  \hat{u}_{\hat{y}|i})_{|\hat{y}_s}$ indicates agreement with respect to $\hat{y}_s$. This procedure is formally presented in Algorithm~\ref{alg:relevantPCBP}.
\begin{table}[t!]
\centering
\caption{Estimation  of class-specific activation ranges ($\alpha$) across all examples for both CapsNet architectures, demonstrating the selection process for the range [min($A_c$), max($A_c$)] for each class. \label{table:classSpecificInter}}
  \begin{adjustbox}{width=0.82\linewidth}
  \begin{tabular}{ccccc}  
    \toprule
    \cmidrule(r){1-5}
&  \multicolumn{2}{c}{\textbf{Dynamic Routing (DR)}} & \multicolumn{2}{c}{\textbf{EM Routing (EMR)}} \\
\textbf{Target} & \textbf{MNIST} & \textbf{SVHN} & \textbf{MNIST} & \textbf{SVHN}\\ \hline
\textit{C$0$} & [-0.44, 0.40] & [-0.3, 0.25] & [0.1, 0.98] & [0.15, 1] \\ 
\textit{C$1$} & [-0.24, 0.3] & [-0.6, 0.6] & [0.12, 1] & [0.15, 0.97] \\
\textit{C$2$} & [-0.2, 0.3] & [-0.25, 0.28] & [0.1, 0.98] & [0.08, 1] \\ 
\textit{C$3$} & [-0.28, 0.3] & [-0.33, 0.29] & [0.1, 1] & [0.1, 1]\\ 
\textit{C$4$} & [-0.33, 0.33] & [-0.6, 0.58]& [0.1, 1] & [0.1, 1] \\ 
\textit{C$5$} & [-0.55, 0.5] & [-0.3, 0.3] & [0.12, 1] & [0.11, 0.99]\\ 
\textit{C$6$} & [-0.1, 0.47] & [-0.5, 0.57] & [0.12, 1] & [0.1, 1]\\ 
\textit{C$7$} & [-0.3, 0.3] & [-0.25, 0.3]& [0.1, 1] & [0.1, 1] \\ 
\textit{C$8$} & [-0.45, 0.22] & [-0.2, 0.25]& [0.1, 1] & [0.06, 0.99] \\ 
\textit{C$9$} & [-0.27, 0.29] & [-0.4, 0.4] & [0.1, 1] & [0.1, 0.99]\\ \hline 
Min/Max among Classes & [-0.55, 0.5] & [-0.6, 0.6] & [0.1, 1] & [0.06, 1]\\ 
$\alpha$  & [-0.55, 0.5] & [-0.6, 0.6] & [0.1, 1] & [0.1, 1]\\
\end{tabular}
\end{adjustbox} 
\vspace*{-0.07in}
\end{table}
\section{Experimental Setup}
\label{sec:experimentalsetup}
\textbf{Datasets:}
Following the settings of well established efforts~\cite{SF17,GT20,RS22,NT20,SM18,JL20,sprj22,patrick2019capsule,MKGS23} we validate our methodology on the MNIST, SVHN, and CIFAR{-}10 datasets. In addition, we include the more complex  CelebA~\cite{ZP15}, and CelebAMask-HQ~\cite{LLW20} (referred to as "CPS") datasets.
MNIST is a grayscale dataset depicting hand-drawn digits. It is composed of 60k training images, from which 10k have been selected for validation, and 10k for testing. SVHN is an RGB dataset depicting number plates. It consists of 63k images for training, 10k for validation, and 26k for testing. 
The CIFAR{-}10 dataset is composed of 60k images covering 10 classes. The dataset has been split into 40k images for training and 10k images for both the validation and the test phases. 
CelebA is composed by 202k facial images consisting of two classes, \textit{male} and \textit{female}. For training the CapsNet, we follow the provided partitioning. CPS is a subset of CelebA with part-level annotations, 2842 test images were used.\\
It is worth mentioning that this selection of datasets differs from the typical selection of tiny and simpler datasets done in existing work~\cite{SF17,GT20,RS22,NT20,SM18,JL20}.

\textbf{Implementation Details: } Our experiments were conducted on CapsNet architectures from  \cite{SF17} and CapsNetEM~\cite{HSF18}. Both were trained on the MNIST and SVHN datasets for 50/32 epochs with a batch size of 32. The input size was set to $28{\times}28$ and $32{\times}32$ for the MNIST and SVHN datasets, respectively. 
On the CIFAR{-}10 dataset, the models were trained for 35/120 epochs with a batch size of 32 on both architectures respectively. In this case, the input size is $64{\times}64$.
Similar to SVHN, CelebA and CPS were resized to an input size of $32{\times}32$. The CapsNet was trained for 7 epochs with a batch size of 32.

In the training step, no data augmentation or hyperparameter tuning were applied and the batch size was determined empirically. Due to limited computational resources, the number of intermediate units per layer in the case of CapsNetEM is reduced from $32$ units to $12$ units, enabling us to train the network on CIFAR{-}10.
For the case of CapsNetEM training configurations, the learning rate was set to \(3 \times 10^{-3}\), allowing for effective convergence during training. We employed a weight decay of \(2 \times 10^{-7}\) to prevent overfitting.
Our experiments were implemented in PyTorch and conducted with NVIDIA GeForce GTX 1060 and DGX-2. We use the adam optimizer \cite{kingma2017adam} with the default learning rate, dropout, and weight decay. In the routing algorithm, {3} routing iterations were used in both CapsNet architectures. The reconstruction loss function employed is the capsule loss which was proposed by \cite{SF17} to ensure effective training. Regarding the number of units in each layer, we used the default ones proposed by both \cite{SF17,HSF18}. To evaluate model performance, we assessed accuracy on the train, validation, and test sets.
\section{Experiments}
\label{sec:results}
\textbf{Classification. }
Table~\ref{tab:performanceDataset} reports the mean classification accuracy of the trained CapsNet on the considered datasets.
In addition, we report the performance from sparser versions of the models based on the identified paths (Sec.~\ref{sec:layerwiseOcclusion}). More specifically, when only the relevant activations were used to make the prediction. 
In the case of the forward path, the relevant units were identified without considering the target classes.
Worth noting is that while the performance using the dense model is on a par with that reported in previous works, the performance using the identified paths is lower. This is more critical for the backward path where it is much more reduced in comparison. It is also noticeable in the results from CIFAR{-}10 that the performance using the identified paths is higher.

After the training, in Fig.~\ref{fig:perWRTRelatedMethods}, we present qualitative examples of reconstructions from MNIST and SVHN datasets trained on a CapsNet with three routing iterations. The first row shows the input images ($s_x$) and the rest of the rows exhibit the reconstruction images (prediction). The figure compares reconstructions generated by the original CapsNet architectures proposed by \cite{SF17,HSF18} with those from two enhanced CapsNets introduced by \cite{MSC21,liu2024capsule}. These reconstruction results are considerably robust while preserving their representative details. In this step, we observe that the most active capsule in the \textit{CC} layer contains sufficient information to reconstruct the original input ($x_s$)

\subsection{Perturbation Analysis}
\label{sec:result-FeatureAnalysisRelevantPerturbation}

This analysis is conducted on the MNIST and SVHN datasets. Following Sec.~\ref{sec:featureAnalysisRelevantPerturbation}, we estimate the empirical activation range $\alpha$ for the $v_j$ vector from the considered datasets. Then, a step size $\xi$ is computed, in order to partition the $\alpha$ range into 12 perturbation possibilities per dimension; similar as in \cite{SF17,NT20,SM18}. We replace the value of each dimension in $v_j$ producing the perturbed vector $v'_{j}$. This vector is pushed through the \textit{decoder} in order to produce reconstructions $\hat{x_s}$ for each perturbed $v'_{j}$.
These reconstructions are then pushed through the CapsNet and classification performance is computed.

Since the estimated ranges might differ from class to class, we defined an overarching range $\alpha$ that encompasses the widest range across all classes. 
Table~\ref{table:classSpecificInter} presents the activation ranges derived from the output vector $v_j$ for both datasets across all classes. It also shows the defined interval $\alpha$, determined by examining the min/max values across all classes.
This is shown in details in Table~\ref{table:classSpecificInter},
for the CapsNet architecture, the activation ranges vary significantly by class (C0 to C9), with C5 exhibiting the widest range on the MNIST dataset [-0.55, 0.5], while C1 and C4 have broader ranges in the SVHN dataset. 
In contrast, the EM Routing (EMR) architecture demonstrates consistent activation ranges close to [0.1, 1] across all classes. The overarching range covers the entire space produced by the training examples reflects the maximum observed activation variations ([-0.55,0.5] for MNIST and [-0.6,0.6] for SVHN). This will help paint a more complete picture of plausible behaviors of the units under study. Moreover, it will ease the comparison w.r.t. the heuristic approach.  
\begin{figure*}[b!]
  \includegraphics[width=0.92\textwidth]{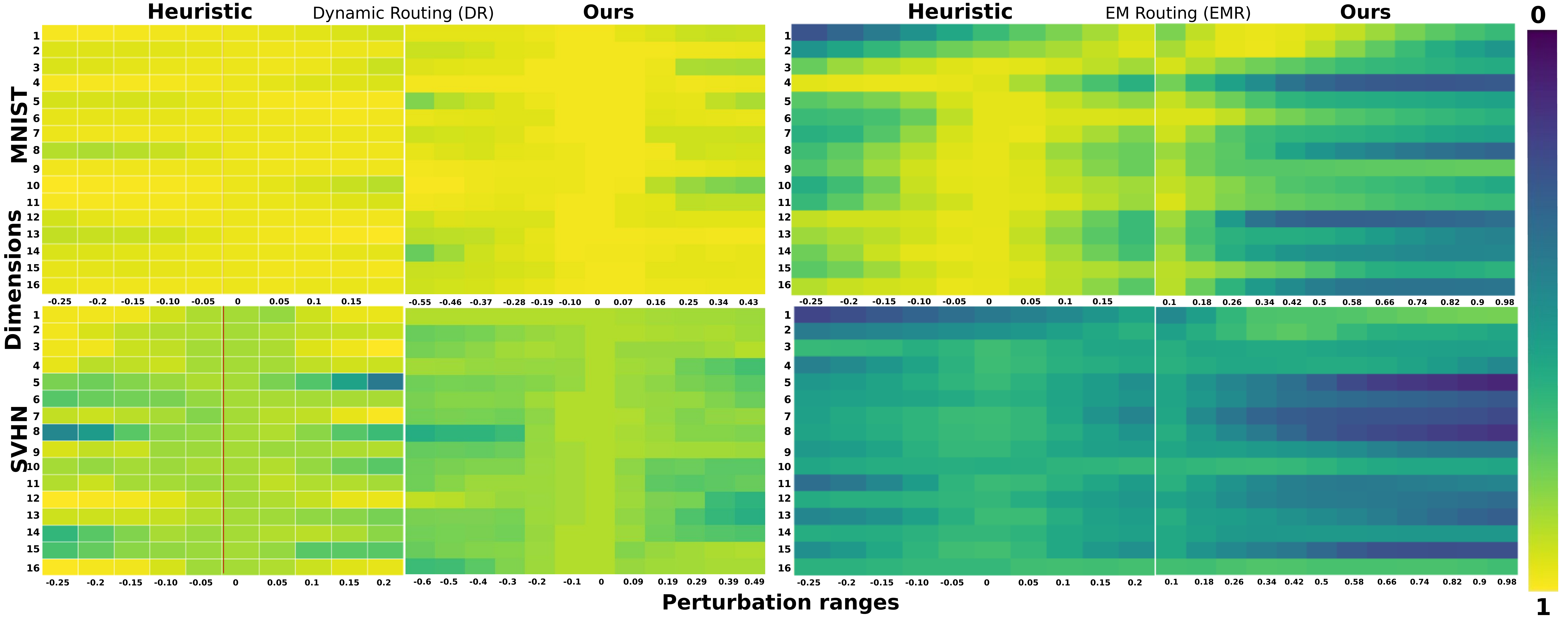}
  \vspace*{-0.12in}
  \caption{Mean classification Accuracy as perturbations are applied to the 16 dimensions of vector $v_j$ on both CapsNet architectures based on the overarching ranges across classes.} \label{fig:histograms_Exper}
  \vspace*{-0.12in}
\end{figure*}

\textbf{Results:} 
As shown in Fig.~\ref{fig:pertubation_Expr}, we test different types of perturbations on the vector $v_j$, with the perturbation range based on the estimated $\alpha$.
Fig.~\ref{fig:pertubation_Expr} (left) presents some qualitative results in the form of reconstructions obtained by the CapsNet architecture from \cite{SF17} by following this procedure. A quick inspection reveals how the applied perturbations effectively provide some insights related to the features encoded by $v_j$.
As can be noted in this figure, each dimension encodes various characteristics of the digits such as thickness, rotation, deformation, and scale. Moreover, the shape (top and bottom regions in some digit classes) of $\hat{x}_s$ shows changes in different forms. 
%
\begin{figure*}[t!]
  \includegraphics[width=0.88\linewidth]{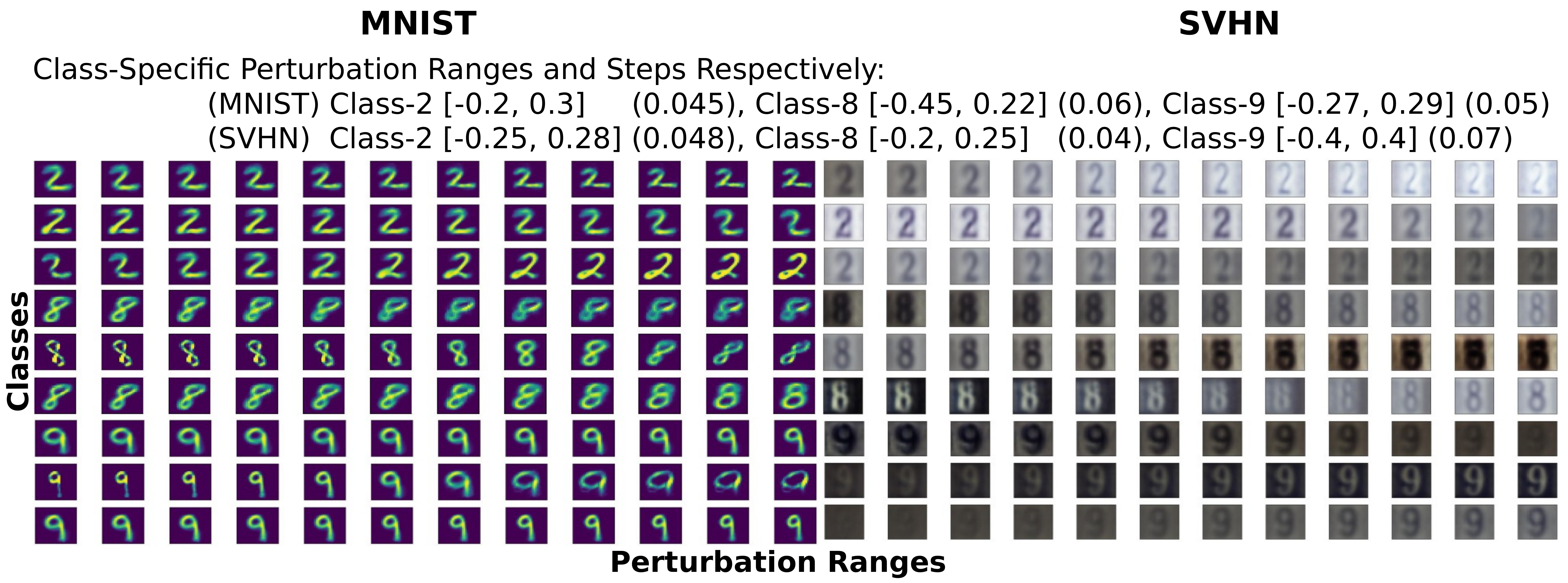}
  \vspace*{-0.10in}
   \caption{Qualitative examples of reconstructed inputs as the vector $v_j$ is perturbed considering Class-Specific intervals ([min($A_c$), max($A_c$)] as defined in Section 4.1) (Classes 2,8, and 9 including the ranges and steps were used)~by~\cite{SF17}. \label{fig:pertAgSpec}}
\end{figure*}
It is clear that changes in specific elements of this feature space ($v_j$) are not exclusive to a single visual feature. As can be noted in Fig.~\ref{fig:pertubation_Expr} (left) (MNIST digits 1-4), perturbations along a single dimension of $v_j$ introduce changes in both rotation and thickness of the generated digit. This suggests some level of feature entanglement.
Moreover, we have observed that a given dimension in the representation space may encode different visual features for different classes. 
Similarly, in Fig.~\ref{fig:pertubation_Expr} (right), we present qualitative results of the reconstructions of the modified $v_j$ from the CapsNetEM model from \cite{HSF18}. 
We observe a similar behavior, where a single dimension seem to encode multiple features across different classes.
In Fig.~\ref{fig:pertubationExprRW2}, we also produce perturbations from the vector $v_j$ of the active capsule, using a perturbation parameter that ranges from $[-0.25, 0.25]$ with 0.05 step size, as defined by \cite{SF17}. These perturbations are applied to both original CapsNet architectures \cite{SF17,HSF18}. 
The top rows show the perturbations from \cite{SF17} and the bottom rows present the perturbations from \cite{HSF18}. We observe that as the perturbation values get larger, the strokes of classes become thicker. We also observe that each dimension encodes various characteristics of the digits such as thickness, rotation, and deformation.
To validate our methodology, we conducted similar experiments using both the heuristic interval and our defined $\alpha$  on two enhanced CapsNets from related works, as shown in Fig.~\ref{fig:pertubationExprRWV1}. 
The results are then compared to the perturbation reconstructions produced by the original CapsNet models across both datasets.
It is evident that the heuristic interval has a limited view/coverage on the characteristics of the perturbed vector $v_j$, especially as the model depth increases. This can be seen in Fig.~\ref{fig:pertubationExprRWV1} (left), where changes are barely noticeable. However, when perturbations are applied using our defined interval $\alpha$  (Fig.~\ref{fig:pertubationExprRWV1} (right)), the reconstructions exhibit different characteristics after the perturbation is done, such as rotation in classes {1} and {6} and deformation in class {9}. Moreover, we still observe similar behaviors where each dimension encodes various characteristics.
These observations are sufficient to conclude that CapsNets might be interpretable but the representations they encode are not disentangled.
\begin{figure*}[b!]
\centering
\includegraphics[width=0.88\linewidth]{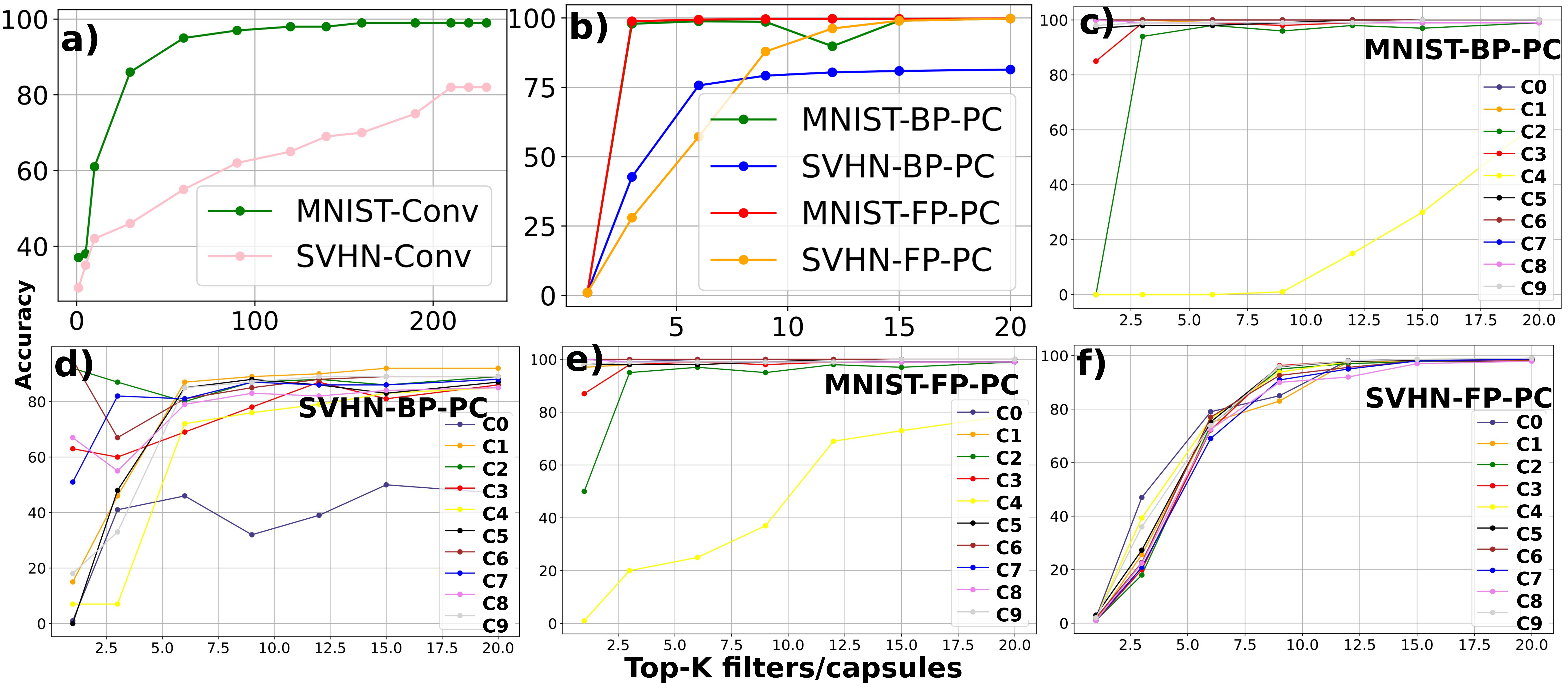}
\vspace*{-0.10in}
\caption{Changes in classification performance as the number of selected relevant units $k$ changes (starting with $k${=} {1}) across CapsNet~\cite{SF17} layers where the relevant units $k$ increased gradually. This is applied on each layer of the network. We also conducted this experiment per class, {C0} to {C9} refer to the classes. \label{fig:topk_performance_mnist_shvn}}
\end{figure*}

To complement the qualitative analysis, Fig.~\ref{fig:histograms_Exper} presents the classification performance that is obtained when each of the mentioned perturbations is applied based on the Heuristic defined in \cite{SF17} and our proposed $\alpha$ interval.
We show the output when the typical (heuristic by \cite{SF17, HSF18}) perturbation is applied, and the output of our perturbation analysis for both the CapsNet and CapsNetEM models.
As can be noted in this figure, overall performance is relatively high in the case of the CapsNet from \cite{SF17} (Fig.~\ref{fig:histograms_Exper}, left). 
A distinct trend is visible in CapsNetEM from \cite{HSF18} (Fig.~\ref{fig:histograms_Exper}, right). 
As can be observed, the performance decreases gradually when modifying $v_j$ based on the maximum value of $\alpha$ ('Ours').
The discrepancies between the standard(heuristic) approach and our method indicate that the activation range to be considered tends to be different than the usual one arbitrarily used in previous works~\cite{SF17,SM18,BA20}.
This observation further stresses that the range used for the analysis (via the $\alpha$ parameter in our case) should be properly estimated.

\noindent\textbf{Class-Specific Perturbation Analysis.} We repeated our qualitative analysis considering class-specific ranges that are defined in Table~\ref{table:classSpecificInter}. As reported in this table, we obtained the intervals for each class on both datasets individually as explained in Section 4.1. Then, we consider the $[min, max]$ as an individual interval per class. 
These intervals are used to produce reconstructions as presented in Fig.~\ref{fig:pertAgSpec}. This figure shows reconstructions as the vector $v_j$ is perturbed based on the defined ranges of specific classes explained earlier ($2$,$8$, and $9$) for both datasets.
As can be seen in Fig.~\ref{fig:pertAgSpec} (MNIST classes 2-8), perturbations within a given dimension of $v_j$ show some joint changes in rotation and thickness, in some cases, and pixel intensity and shape, in others. 
That is, modifying a single unit leads to the modification of multiple visual features.

\subsection{Layer-Wise Relevant Unit Selection}
\label{sec:result-LayerWiseOcclusion}
\begin{figure*}[b!]
\centering
\includegraphics[width=0.88\linewidth]{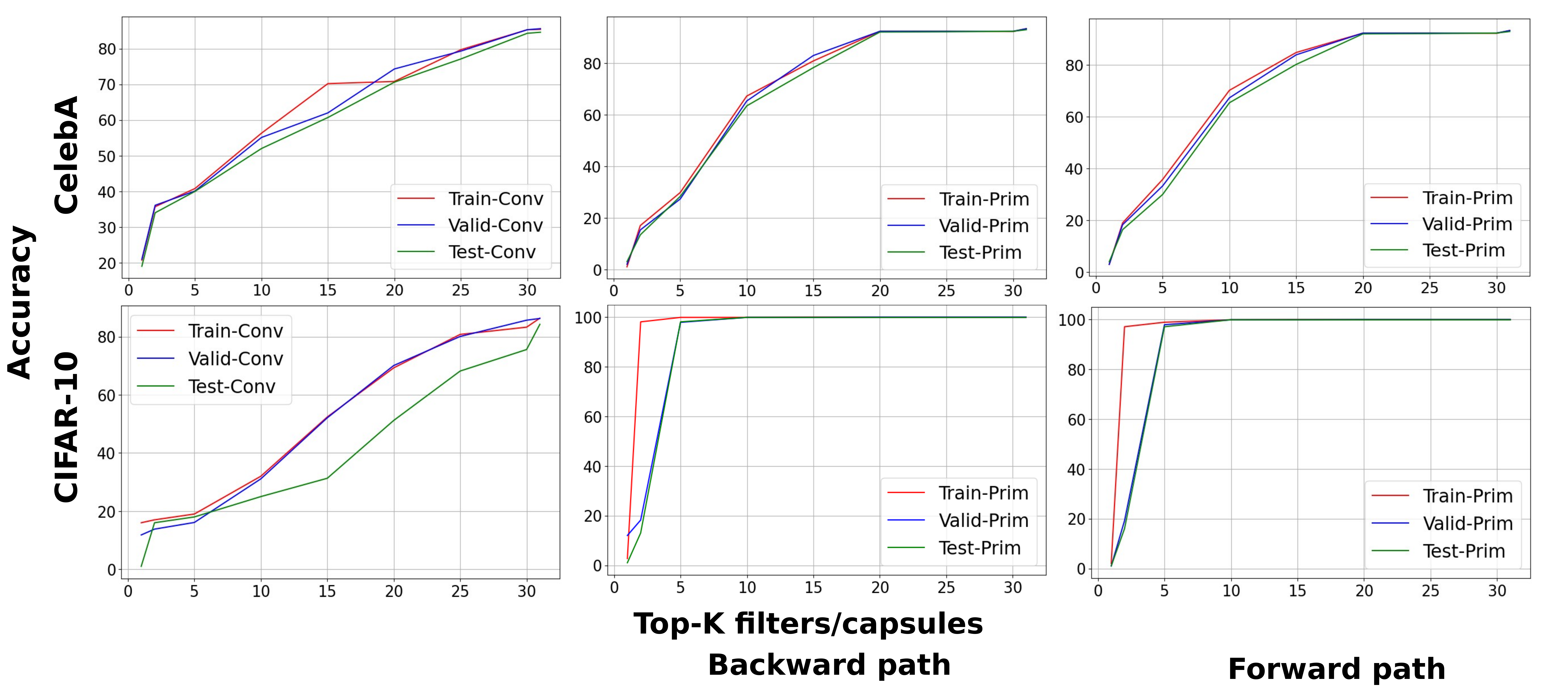}
\vspace*{-0.11in}
\caption{Changes in classification performance as the number of selected relevant units $k$ changes (starting with $k${=} {1}) across the layers from DR CapsNet~\cite{SF17}. This is applied to each layer of the network on the CelebA and CIFAR-{10} datasets (both forward and backward paths are considered for the case of capsule layers). \label{fig:topkPerfCelebCifar}}
\end{figure*}

This experiment is also conducted on the MNIST and SVHN datasets. Following the methodology introduced earlier, we define the \textit{forward (FP)} and \textit{backward (BP)} activation paths based on the top-$k$ units on the CapsNet.
The $k$ values used in these experiments were determined empirically. Then, the final values were defined based on the classification performance presented in Fig.\ref{fig:topk_performance_mnist_shvn}.
The cross-validation procedure led to the selection of $k{=}35$ filters for the \textit{Conv} layer and $k{=}10$ for the \textit{PC} layer for both datasets. 
In Fig.~\ref{fig:topk_performance_mnist_shvn}, we report test classification performance as the number ($k$) of selected units is increased for CapsNet by \cite{SF17}.
We report detailed results of the impact of modifying different layers of the CapsNet. Specifically, we report results when only the \textit{Conv} layer is modified (Fig.~\ref{fig:topk_performance_mnist_shvn}.a and Fig.~\ref{fig:topkPerfCelebCifar} ($1^{st}$ column), when only the \textit{PC} layer is modified (Fig.~\ref{fig:topk_performance_mnist_shvn}.b and Fig.~\ref{fig:topkPerfCelebCifar} ($2^{nd}$ \& $3^{rd}$ column) and when both layers are modified at the same time, i.e. when the complete path is considered per class, for the MNIST (Fig.~\ref{fig:topk_performance_mnist_shvn}.c \& e ) and SVHN (Fig.~\ref{fig:topk_performance_mnist_shvn}.d \& f) datasets. We can notice that the performance improves as the number of $k$ units increases.\\
\begin{figure*}[t!]
\centering
\includegraphics[width=0.90\linewidth]{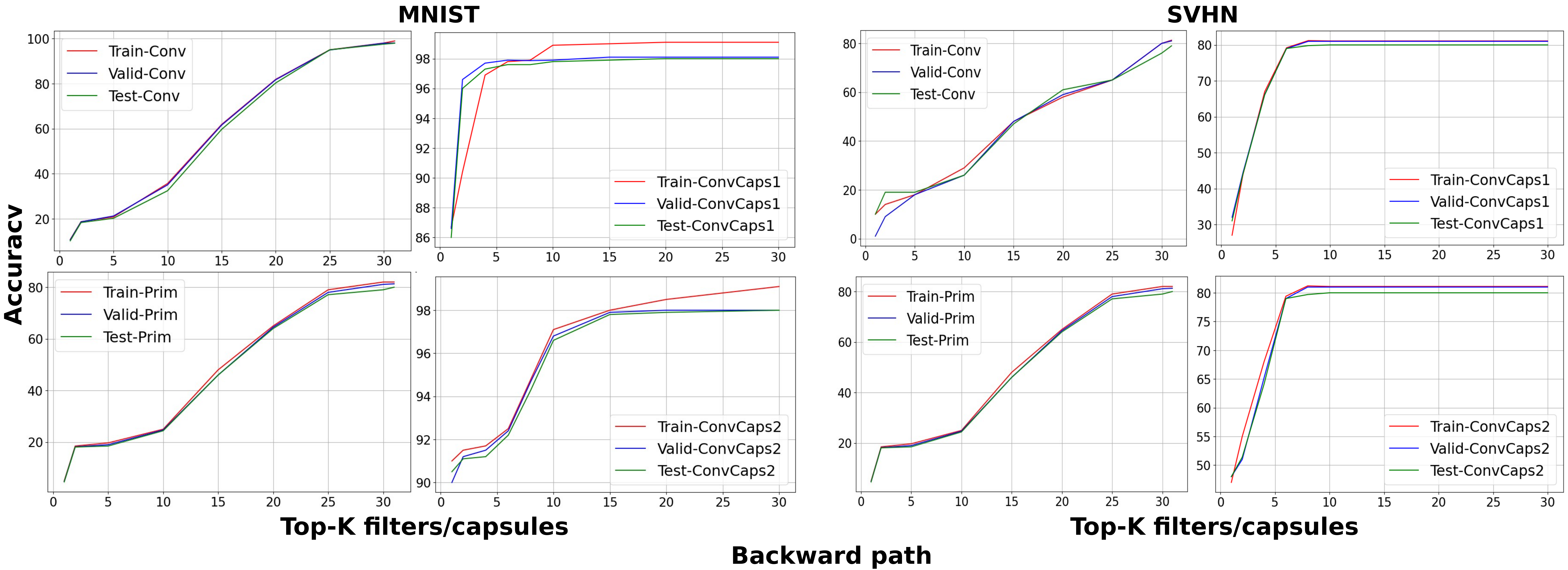}
\vspace*{-0.10in}
\caption{Changes in classification performance as the number of selected relevant units $k$ changes across \textit{CapsNetEM}~\cite{HSF18} layers where the relevant units $k$ increased gradually. The \textit{backward} path estimation is used to evaluate performance impact, where $k$ values are incremented, indicating how selecting more relevant units influences the network’s ability to classify accurately across layers on both MNIST and SVHN datasets. \label{fig:classfPerforEM}}
\end{figure*}
\begin{figure*}[b!]
\centering
\includegraphics[width=0.90\linewidth]{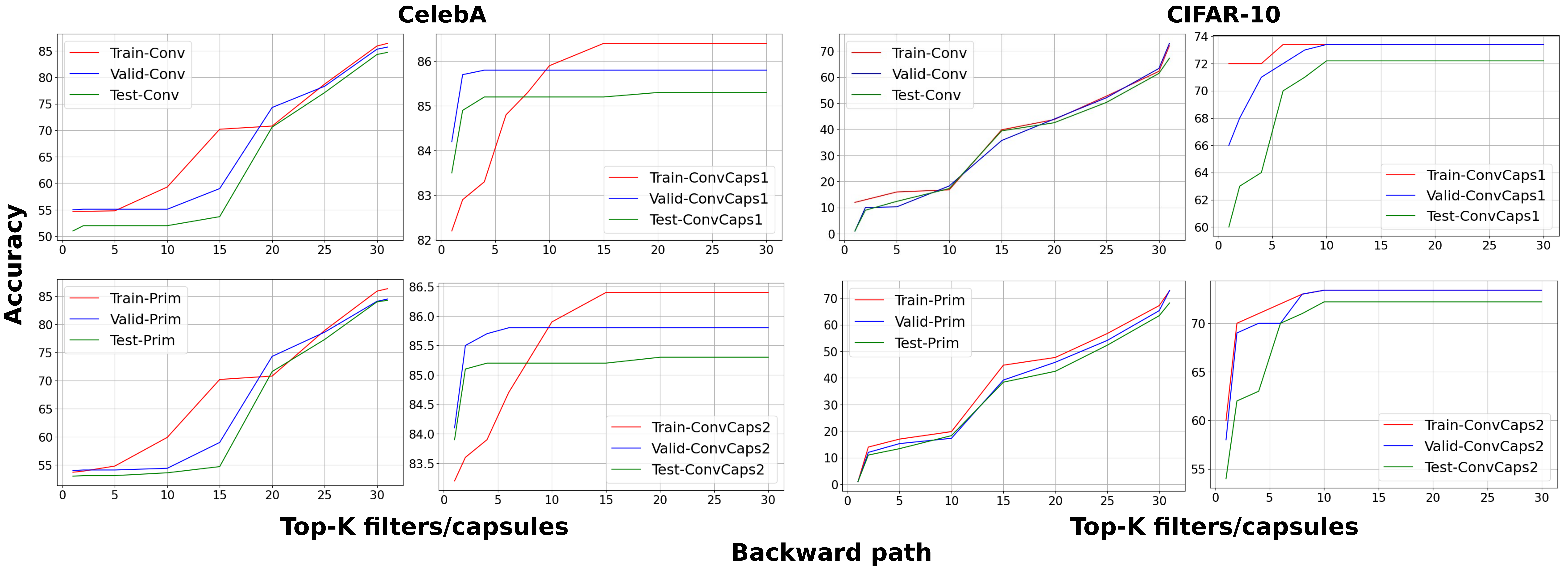}
\vspace*{-0.10in}
\caption{Changes in classification performance as the number of selected relevant units $k$ changes across \textit{CapsNetEM}~\cite{HSF18} layers where the relevant units $k$ increased gradually on both CelebA and CIFAR{-}10 datasets (\textit{backward} path estimation). Number $k$ units was initially set to 1. \label{fig:classfPerEMForPass2}}
\end{figure*}
\begin{figure*}[b!]
\centering
\includegraphics[width=0.90\linewidth]{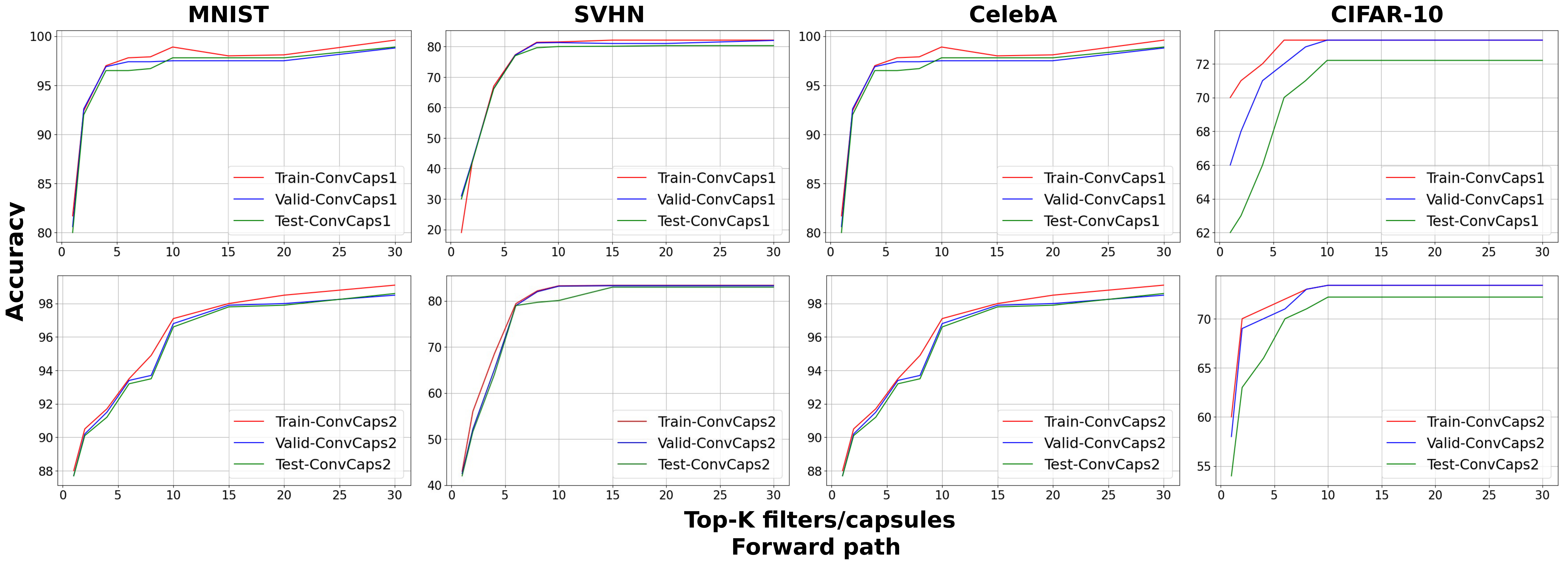}
\vspace*{-0.10in}
\caption{Changes in classification performance as the number of selected relevant units $k$ changes across the layers from EM \textit{CapsNetEM}~\cite{HSF18}.
The top-{k} units are selected based on the \textit{forward} path estimation method.\label{fig:classfPerEMForPass}} 
\end{figure*}
\begin{figure}[t!]
\centering
\includegraphics[width=\linewidth]{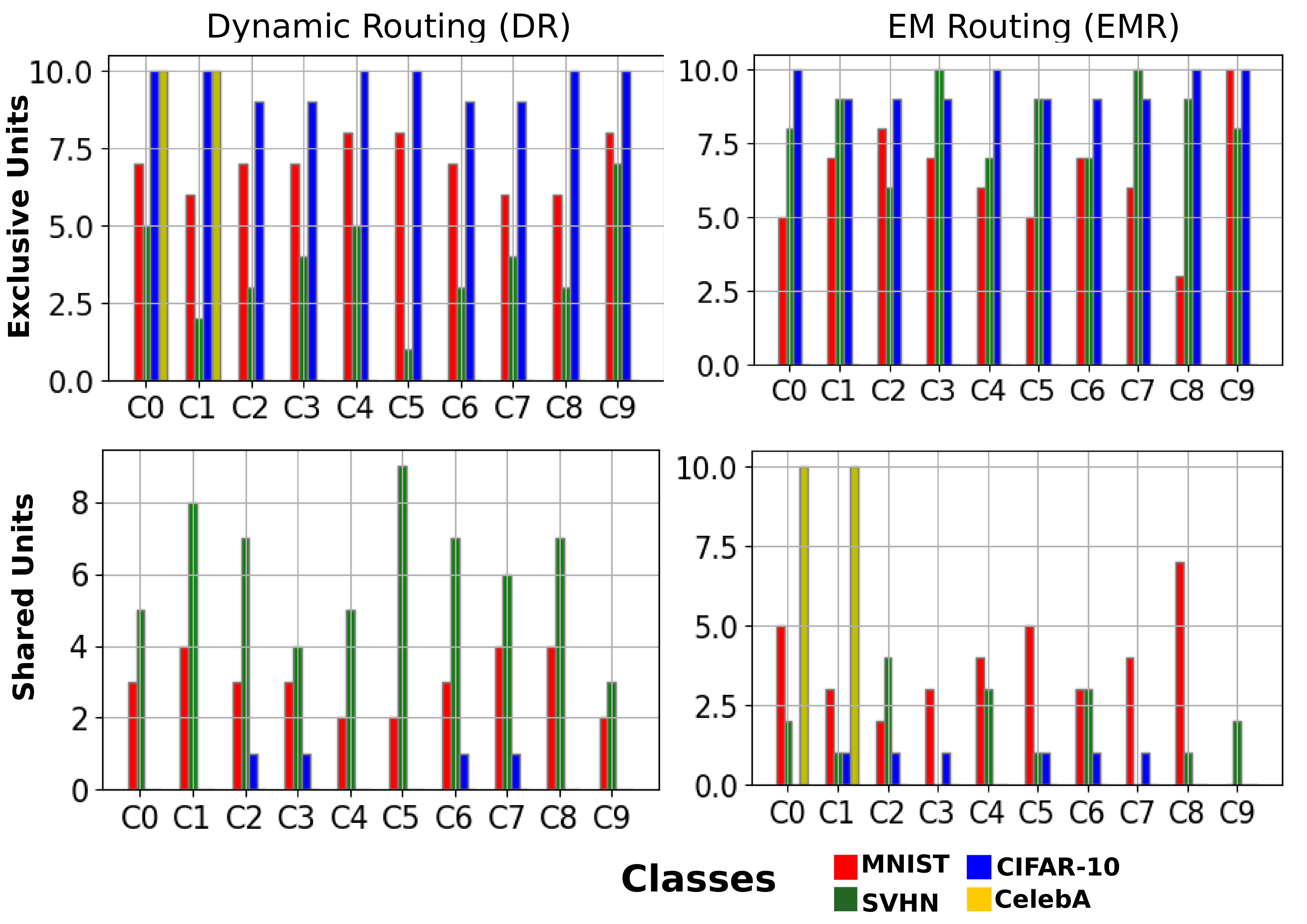}
\vspace*{-0.25in}
\caption{Analysis of the number of class-exclusive for each class and shared top-$k$ units across CapsNet Architectures: A comparative study of dynamic routing (DR) and EM routing (EMR).\label{fig:sharedUnit}}
\vspace*{-0.14in}
\end{figure}
\textbf{Convolutional Layer Ablation.} 
When only the \textit{Conv} layer is modified, as shown in Fig.~\ref{fig:topk_performance_mnist_shvn}.a, we notice that performance on the test set is quite acceptable for the case of MNIST. The story is different on SVHN, where there is a drop in performance when only considering the selected ($k{=}35$) units. This suggests a significant domain shift between the validation and test subsets of that dataset.

\textbf{Capsule Layer Ablation.} 
When only the \textit{PC} layer is modified as presented in Fig.~\ref{fig:topk_performance_mnist_shvn}.b, we notice that performance is relatively high in most cases. For the selected units ($k{=}10$), we notice that the units selected via the \textit{forward (FP)} path achieve higher performance than their \textit{backward (BP)} counterparts on both datasets.\\
Additional experiments on the CapsNetEM architecture show that the outcomes exhibit trends similar to those of CapsNet~\cite{SF17}.
On the one hand, Fig.~\ref{fig:classfPerforEM} and Fig.~\ref{fig:classfPerEMForPass2} present the changes in classification performance as the number of selected relevant units $k$ increases gradually for the \textit{backward} path estimation across the used datasets.
On the other hand, Fig.~\ref{fig:classfPerEMForPass} shows the corresponding results for the (\textit{forward})) path. 
In both path estimation methods, we observe a similar pattern: starting with a very small value of $k$, the classification performance gradually improves as the number of $k$ units increases.
Based on the results presented above, we observe that a small number of relevant $k$ units is sufficient to define the optimal path through the network. Therefore, utilizing all units within the network is unnecessary.

\begin{table}[t!]
\centering
  \caption{Number (\%) of shared top-$k$ units among two classes on CapsNets DR by~\cite{SF17}  and EMR by~\cite{HSF18} (DR: dynamic routing and EMR: EM routing).\label{tab:OverlappingTwoClasses}}
  \vspace*{-0.08in}
  \begin{adjustbox}{width=\linewidth}
  \begin{tabular}{lllllllll}
    \toprule
    \cmidrule(r){1-9}
     CapsNet Type & Dataset & C1-C7 & C2-C8 & C3-C5 & C4-C8 & C6-C8 & C7-C9 & C0-C9\\
    \midrule
     Dynamic Routing & MNIST & 4 (40) & 2 (20) & 3 (33) & 2 (20) & 1 (10) & 2 (20) & 1 (10)\\
     (DR) & SVHN & 4 (40) & 2 (20) & 5 (50) & 1 (10) & 4 (40) & 2 (20) & 2 (20)\\
    \midrule
    \multicolumn{2}{c}{}& C1-C6 & C3-C8 & C4-C5 & C0-C8 & C5-C8 & C5-C7 & C6-C9\\
     \midrule
     EM Routing& MNIST & 2 (20) & 3 (30) & 1 (10) & 3 (30) & 3 (30) & 3 (30) & 0 (0)\\
     (EMR) & SVHN & 0 (0) & 0 (0) & 1 (10) & 1 (10) & 0 (0) & 0 (0) & 1 (10)\\
  \end{tabular}
  \end{adjustbox}
  \vspace*{-0.10in}
\end{table}
We looked for shared and exclusive units on both CapsNets models among the top-$k$ units selected for each class by our layer-wise relevant unit selection method (Sec.~\ref{sec:layerwiseOcclusion}).
Fig.~\ref{fig:sharedUnit} presents the number of top-$k$ units exclusive for each class individually and those shared among all the other classes on both architectures. 
The figure shows how some units remain class-specific, meaning they activate strongly for a particular class while remaining inactive for others, thereby serving as distinct identifiers for that class. 
In contrast, shared units, which are activated across several classes, contribute to general feature representations that can be beneficial for modelling similarities across classes.\\
In MNIST, we noted that classes with similar appearance had a higher number of shared units. In particular, classes [1,7], [3,5], [3,8], [8,9], with 4, 3, and 3 shared units, respectively. This is also shown in Table~\ref{tab:OverlappingTwoClasses}.
In SVHN, on the quantitative side, we noted a higher number of shared units. On the qualitative side, the relation between class and the relevant unit was less pronounced, possibly due to the occurrence of parts of other digit instances co-occurring in the input images. In this dataset, classes [0,2], [5,9], [6,8], [2,5], [3,4] had 6, 6, 5, and 5, shared units, respectively. 
For the CIFAR{-}10 dataset, the classes bird-cat were the ones that had the highest number (2) of shared units.
Worth mentioning is that it was observed across all the datasets, that the activation magnitude of the shared units was significantly low in comparison with the units exclusive to each class.
For CelebA/CPS, all units were exclusive between class male and female.
\begin{figure*}[t!]
   \centering
  \includegraphics[width=0.88\linewidth]{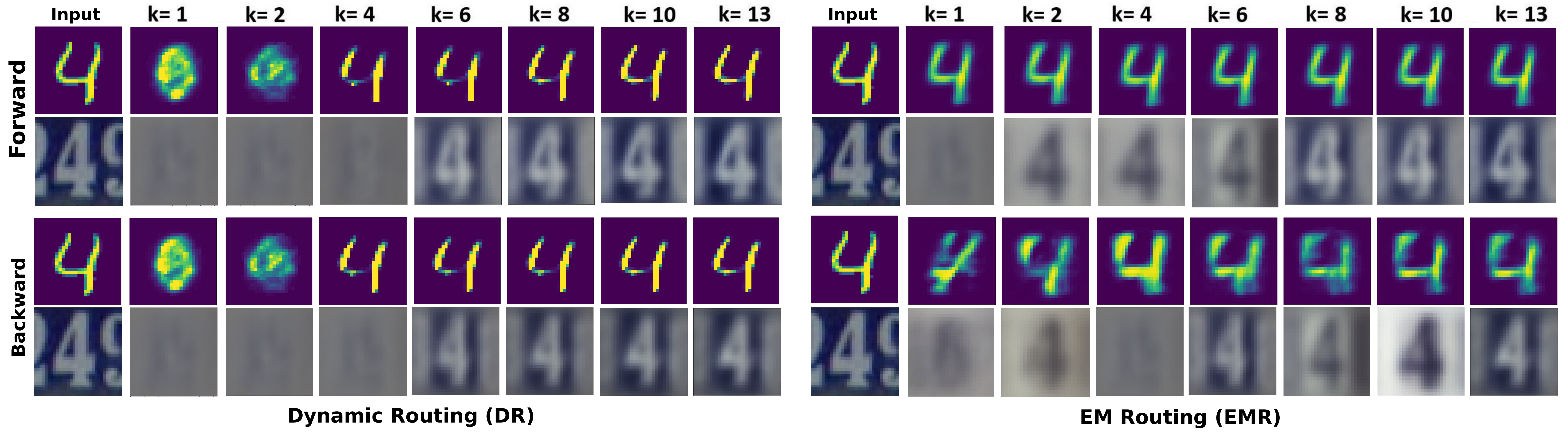}
  \vspace*{-0.10in}
   \caption{Reconstructed input images from the MNIST and SVHN dataset, where reconstructions are generated by selectively using only a small number  of $k$ units from the \textit{PC} layer. The left side illustrates reconstructions produced with DR CapsNet~\cite{SF17}, and the right side shows those from EMCapsNet~\cite{HSF18}. Notably, the EMCapsNet reconstructions exhibit greater sharpness and detail, suggesting that the EMCapsNet may capture more precise spatial and feature-based information within limited relevant units.\label{fig:PathTopkDR_EM}}
\end{figure*}

In the case of CapsNetEM, Table.~\ref{tab:OverlappingTwoClasses} (bottom) shows fewer unit shared among two classes in the case of MNIST e.g., classes [6,9], [4,5] had 0 and 1 units respectively.
We observed a higher number of shared units among certain classes, namely [2, 8], [5, 8], and [5, 7], each having 3 shared units. However, in the case of SVHN, the majority of units are exclusive. 
Fig.~\ref{fig:sharedUnit} (bottom) illustrates that classes with similar appearances tend to have a higher number of shared units.

These are also observed on the qualitative results presented on Fig.~\ref{fig:PathTopkDR_EM} (left) and Fig.~\ref{fig:PathTopkDR_EM} (right). 
In this figure, we present the reconstructed inputs when only a small amount of units $k$ in the PC layer are propagated through the network. The results are shown for paths estimated in both a \textit{forward} (top) and a\textit{backward} (bottom) manner. A clear difference in sharpness that can be observed between the reconstructions when $k=1$ and when $k=6$ is propagated, for both datasets.
There it can be observed that, for both datasets, the quality of the reconstructed images stabilizes around the selection of $k{=}10$ units. Moreover, it is noticeable that, for the case of SVHN, reconstructions produced from units in the \textit{forward} path are sharper that those from the \textit{backward} path.
These observations support the difference in performance across paths observed in Table~\ref{tab:performanceDataset}.
These observations lead to the conclusion that utilizing only the relevant $k$ units in the \textit{PC} layer in both architectures effectively defines the activation paths for reconstruction. By propagating only these units, we can evaluate the model's ability to reconstruct the input images, 
demonstrating that the network can maintain performance with minimal/focused activations.
\begin{figure*}[b!]
   \centering
  \includegraphics[width=0.88\linewidth]{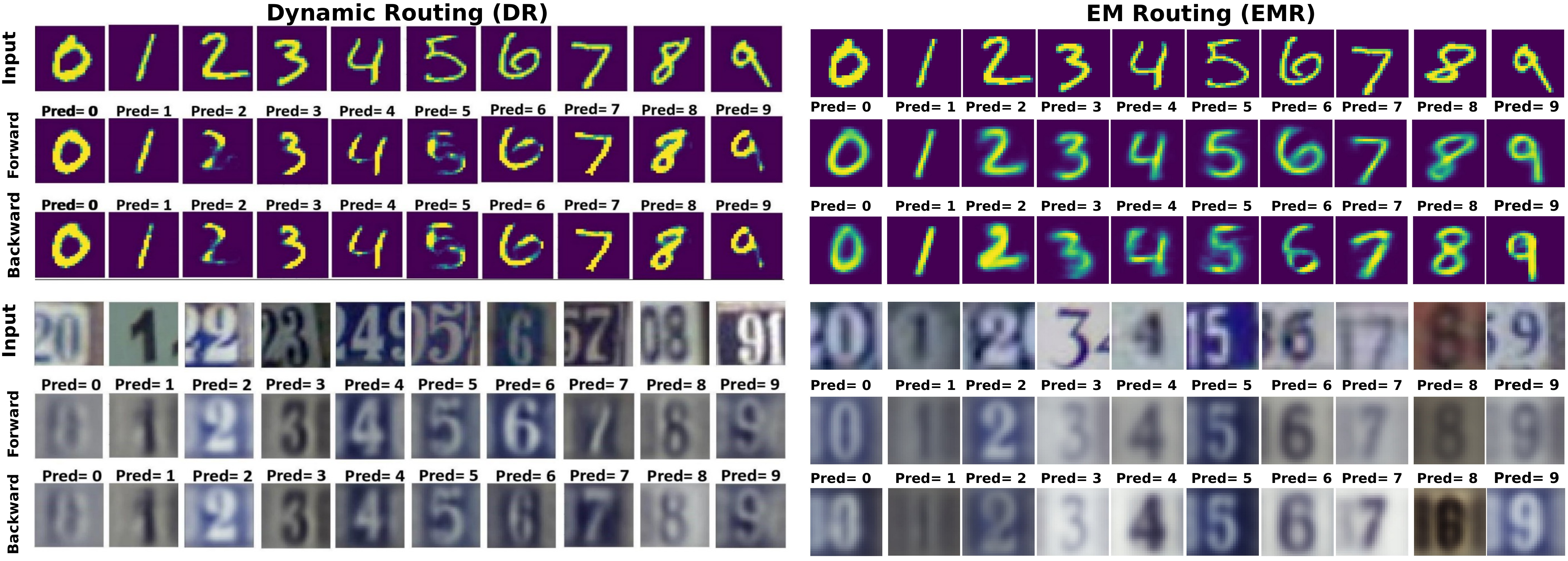}
  \vspace*{-0.10in}
   \caption{Reconstructed input images are produced by considering only relevant units in the network, specifically $k{=}35$ units from the Conv and $k{=}10$ units from the PC layer. Reconstructions on the left are produced using the DR CapsNet~\cite{SF17}, and those on the right use the updated EMCapsNet~\cite{HSF18}. This comparison highlights how each architecture reconstructs inputs when constrained to a relevant set of active units. Differences in reconstruction quality and detail show the impact of each network’s routing mechanism and unit selection on the quality of the generated images. \label{fig:relevantPathDR_EM}}
\end{figure*} 

\textbf{Network Ablation.} When the complete path is used, we notice that overall classification performance remains high. By inspecting the per-class performance, we observe that very few cases (digit-4 for MNIST and digit-0 for SVHN) suffer a significant drop when only considering the selected top-k units per layer.
Fig.~\ref{fig:relevantPathDR_EM} (left) shows qualitative reconstructions that were obtained when the selected $k$ relevant units ($k$ {=} $35$ \textit{Conv} filters/ $k$ {=} $10$ capsules) are only propagated through the architecture \cite{SF17}.
This clearly illustrates the network’s ability to produce high-quality reconstructions with a reduced set of relevant units. By limiting propagation to only these units, we observe that the network not only conserves computational resources but also focuses on the most informative features, which enhances interpretability.
This reduction in active units reinforces the idea that only a subset of filters and capsules contribute meaningfully to reconstruction.
For MNIST, we notice that, while the digit sketches were not complete, the selected regions of the sketches seem to be sufficient to characterize the digit classes. For SVHN, we notice that background information, usually in the form of other digit regions, seem to be suppressed in the reconstructions. This suggests the selected units focus on the foreground objects, i.e. the digit of interest. These observations support our analysis on these relevant units for the search of possibly encoded parts.
Fig.~\ref{fig:relevantPathDR_EM} (right) presents qualitative reconstructions obtained by propagating only the selected $k$ relevant units ($k=35$ \textit{Conv} filters or $k=10$ capsules) through the \textit{CapsNetEM} proposed by \cite{HSF18}.
We anticipated similar behavior using CapsNetEM routing. However, we observed that the units do not seem to be class-specific. Most importantly, as shown in Fig.~\ref{fig:relevantPathDR_EM}, we observe that even with a small number of units, both networks were able to successfully reconstruct and predict the classes correctly.
\subsection{Measuring \textit{Part-Whole} Relationship Encodings}
\label{sec:PartWhole}
This experiment aims to measure the level to which features encoded in a CapsNet encode \textit{part-whole} relationships. 
Towards this goal, inspired by methods proposed for CNNs \cite{BK17,GM18}, we measure the spatial overlap between the internal responses of relevant capsules at layer $l$ (\textit{Parts}) and the internal response at a layer $l+1$ (\textit{Whole}).
In our experiments, we measure this overlap between the \textit{PC}~($h^l$) and \textit{CC}~($h^{l+1}$) layers. The responses of relevant capsules were represented as heatmap $h$~\cite{SV14} computed by, first, estimating the prediction $\hat{y}_s$ produced by input $x_s$ when only the activations of a given unit are propagated forward during inference.
Then, given the prediction $\hat{y}_s$ we compute the gradients of this prediction w.r.t. the input $x_s$.  

The overlap between~$h^l$ and $h^{l+1}$ is measured via the Relevance Mass Accuracy (\textit{RMA}) metric~\cite{AO22}, which provides a value in the range [0,1] indicating the level of overlap. 
The \textit{RMA} requires~$h^{l+1}$ to be a binary matrix. To meet this requirement, we binarized the response~$h^{l+1}$ from \textit{CC}. Each~$h^l$ was normalized in a per-unit basis by considering the min/max values from all the heatmaps related to that unit. We report results for various threshold values ($0.1$, $0.25$ \& $0.5$).
When measuring the overlap between~$h^l$ and $h^{l+1}$, we focus our analysis on the top-$k$ ($k{=}200$) units selected following the \textit{BP} (Sec.~\ref{sec:result-LayerWiseOcclusion}) extraction procedure. 


\textbf{Isolated Unit Analysis. }
Table~\ref{tab:RMAScoresMergedNets}~(DR and EMR / Isolated) presents the mean/STD \textit{RMA} scores of the top{-}1 relevant unit separately across all classes for different thresholds ($0.1$, $0.25$, and $0.5$) per dataset.
For a given input example, overlap is computed for every response pair [$h^l$ (from \textit{PC}), $h^{l+1}$ (from \textit{CC})] produced by each of the selected top-200 units, in isolation.

\begin{figure*}[b!]
   \centering
  \includegraphics[width=0.90\linewidth]{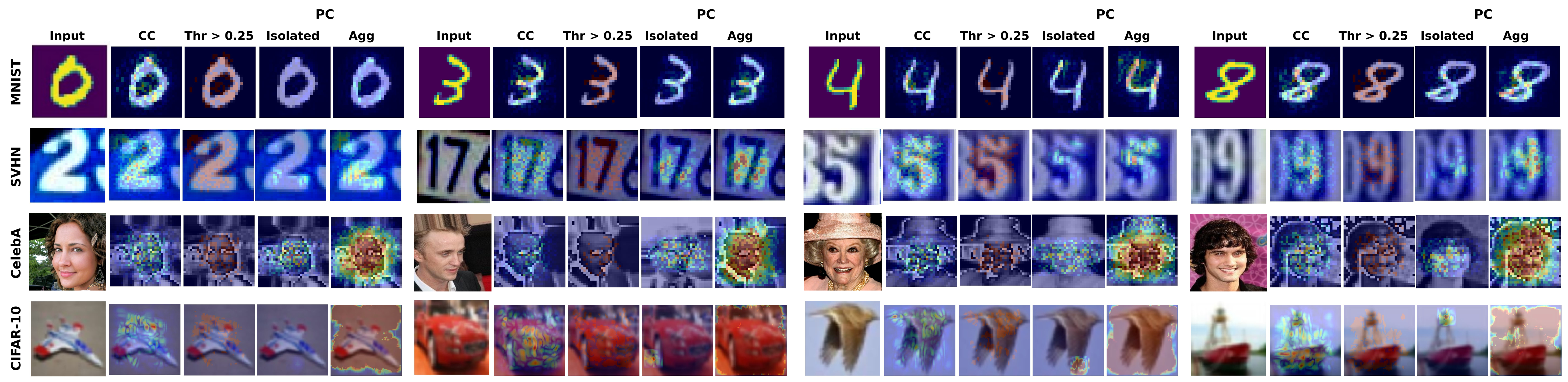}
  \vspace*{-0.10in}
   \caption{Qualitative examples of the responses considered for the computation of relevance mass accuracy (RMA) with $0.25$ threshold~(Thr) applied on \textit{CC}. Heatmaps are generated based on top-$k{=}1$ relevant unit (isolated) and when top-$k{=}200$ (Aggregated)~\cite{SF17}.
  \label{fig:semanticParts}}
\end{figure*}

Across datasets, we can notice that the \textit{RMA} scores significantly drop as the threshold increases.
This is to be expected since a higher threshold enforces a small region in $h^{l+1}$ (\textit{$CC$}), which is harder to cover accurately. However, it is noticeable how, even for more lenient threshold values ($0.1$ and $0.25$) the overlap scores remain relatively low. 
We notice that the scores related to SVHN and CIFAR{-}10 seem to be the highest. Overall, the mean of \textit{RMA} is lower than what was anticipated. This could be due to the inner-workings of the \textit{RMA} metric. 
Fig.~\ref{fig:semanticParts} shows qualitative examples of the responses considered for the computation of the \textit{RMA} with threshold of $0.25$ applied to $CC$ layer heatmaps. The heatmps are generated using CapsNets by \cite{SF17} based on top{-}k{=}$1$ (isolated).
As we can see in this figure (Fig.~\ref{fig:semanticParts}), the $h^{l+1}$ ($CC$) responses are sparse even when binarized with a relatively low threshold value (0.25). We notice a similar behavior on the $h^{l}$ responses from $PC$. 
For instance, as depicted in Fig.~\ref{fig:semanticParts} (4th column), for the class aeroplane from CIFAR{-}10, the $h^{l}$ (\textit{PC}) response is not only sparse, but also covers a very small region.
This affects the overlap score during the estimation. Moreover, the very focused nature of the $PC$ responses, see Fig.~\ref{fig:semanticParts}, suggests, first, that each of the relevant units model very specific details from the input data; and second, that compositions of all the $h^{l}$ responses might achieve higher coverage.

\textbf{Aggregated Unit Analysis.}
Complementing the experiment above, we conduct a second analysis where the overlap is computed between aggregated $h^l$ responses from $PC$ and the single $h^{l+1}$ response they produce at $CC$. This aggregation occurs at the pixel level by taking the maximum response per pixel location across the response maps produced by the considered top-200 relevant units.
\begin{figure*}[t!]
   \centering
  \includegraphics[width=0.90\linewidth]{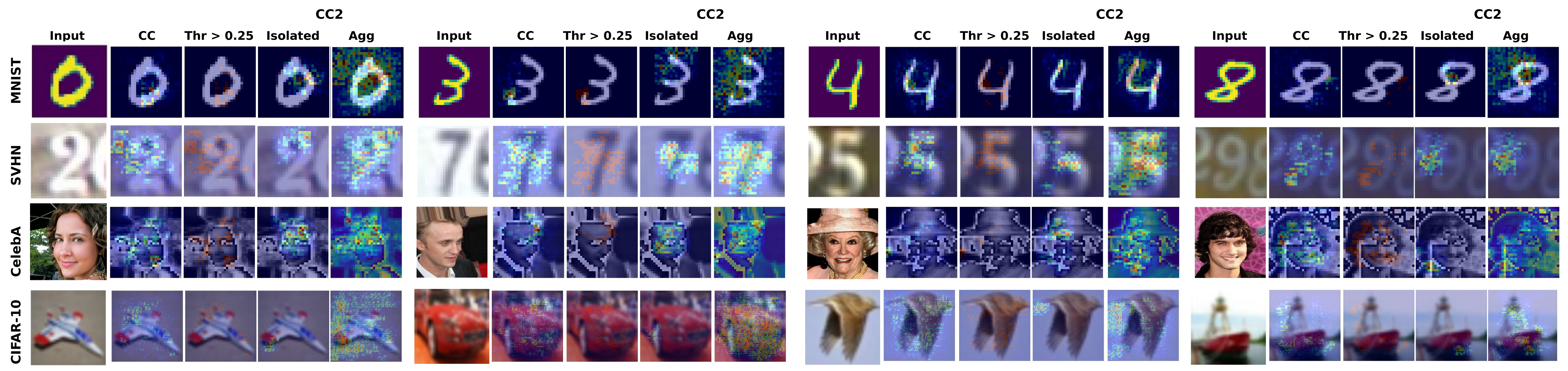}
  \vspace*{-0.10in}
   \caption{Qualitative examples of the responses considered for the computation of relevance mass accuracy (RMA) with $0.25$ threshold~(Thr) applied on \textit{CC}. Heatmaps are generated based on top-$k{=}1$ relevant unit (isolated) and when top-$k{=}200$ (Aggregated)~\cite{HSF18}.
  \label{fig:partWholeEM}}
\end{figure*}
Fig.~\ref{fig:semanticParts} shows qualitative examples of the responses considered for the computation of the \textit{RMA} with threshold of $0.25$ applied to heatmaps from the units of the $CC$ layer. The heatmps are generated using CapsNets by \cite{SF17} based on top{-}k{=}$200$ (aggregated).
A quick inspection of the qualitative results presented in 
Fig.~\ref{fig:semanticParts}~\textit{(Agg)} show two different trends. For the case of CIFAR{-}10 and CelebA (Fig.~\ref{fig:semanticParts}, $2^{nd}$ \& $4^{th}$ column), the effectiveness of the compositions \textit{(Agg)} from different relevant units is evident. Here, the composition covers a significant region belonging to the objects of interest.
On the contrary, for MNIST and SVHN  (Fig.~\ref{fig:semanticParts}, $1^{st}$ \& $3^{rd}$ column), considering additional units does not lead to higher coverage.
These two observations might be pointing at the simplicity of such datasets and the inherent complexity of making predictions on them. More concretely, for more complex cases (CIFAR{-}10 \& CelebA) the units are steered towards learning complementary features; leading to larger coverage when aggregated. For the latter simpler,  more constrained,  case (MNIST \& SVHN) a lower amount of features are needed. This results in more redundancy across the features encoded by the units, and might be the reason behind the reduced coverage from the aggregated responses.

We noticed similar behavior in the case of CapsNetEM, 
We present qualitative examples of the responses considered for the computation of the \textit{RMA} with a threshold of $0.25$ applied to heatmap from units of the $CC$ layer. The heatmps are generated using CapsNets by \cite{SF17} based on top{-}k{=}$1$ (isolated) and top{-}k{=}$200$ (aggregated).
This figure (Fig.~\ref{fig:partWholeEM}) shows similar trends to CapsNet where the units after aggregating show higher coverage.
Table~\ref{tab:RMAScoresMergedNets}~(right) lists the mean \textit{RMA} scores related to this experiment in both CapsNet architectures. It is noticeable that for most of the cases even after aggregating a significant amount of relevant units into $h^l$, the \textit{RMA} score remains low. Moreover, a significant drop is observed in the CIFAR{-}10 and CelebA datasets.
The reason for these low values may find its origin in the relationship between the aggregated response $h^l$ and the binarized response $h^{l+1}$ when considered in the RMA computation.
As indicated earlier, for CIFAR{-}10 and CelebA, a larger-coverage response $h^l$ is observed. This is clearly an over-estimation when compared to the binarized $h^{l+1}$ $(CC)$ that is used as a reference. This is different in the case of MNIST and SVHN, where the compared responses are roughly the same.
In our experiments with CapsNetEM routing, we noticed similar trends where the overlap is relatively low.

While the results presented in Table~\ref{tab:RMAScoresMergedNets} suggest that CapsNets may not effectively encode \textit{part-whole} relationships, it is worth noting that the observed low overlap may have its origin in other sources. More specifically, beyond an unsatisfactory overlap between the responses from $PC$ and $CC$, the observed low RMA scores can also be attributed to the sparsity of the considered responses (see Fig.~\ref{fig:semanticParts} \& Fig.~\ref{fig:partWholeEM}). As shown in~\cite{VandersmissenExpEval}, RMA scores and other metrics for measuring overlap tend to favor smooth continuous heatmaps.
In addition, we currently focused on the analysis of the top-200 relevant units due to the high computational costs required for a fine-grained analysis.\\
\begin{table}[t!]
  \centering
  \caption{RMA scores computed from the top-{200} relevant units identified in each analysis. The left side shows scores when DR CapsNet and EMCapsNet units are evaluated separately, while the right panel displays scores when both methods are aggregated. Each analysis is conducted over multiple thresholds, allowing a consistent comparison on how relevance varies across routing mechanisms. \label{tab:RMAScoresMergedNets}}
  \vspace*{-0.10in}
  \begin{adjustbox}{width=\textwidth}
  \begin{tabular}{l||cccccc||cccccc}
       & \multicolumn{6}{c||}{\textbf{Dynamic Routing (DR)}} & \multicolumn{6}{|c}{\textbf{EM Routing (EMR)}} \\ \toprule
       & \multicolumn{3}{c||}{\textbf{Isolated}} & \multicolumn{3}{|c|}{\textbf{Aggregated}} & \multicolumn{3}{||c||}{\textbf{Isolated}} & \multicolumn{3}{|c}{\textbf{Aggregated}} \\
      \toprule   
    Dataset/Thr & $0.1$ & $0.25$ & $0.5$ & $0.1$ & $0.25$ & $0.5$ & $0.1$ & $0.25$ & $0.5$ & $0.1$ & $0.25$ & $0.5$ \\ \cmidrule(r){1-13}
    
    MNIST & $31\pm6$ & $12\pm$4 & $3\pm$1 & $31\pm5$ & $12\pm3$ & $3\pm4$ & 22$\pm$14 & 8$\pm$5 & 2$\pm$1 &  22$\pm$13 &  7$\pm$5  & 2$\pm$1 \\

    SVHN & $60\pm10$ & $26\pm8$ & $6\pm3$ & $55\pm10$ & $25\pm7$ & $7\pm2$ & 52$\pm$16  & 19$\pm$9 & 4$\pm$3 & 41$\pm$12 & 14$\pm$6 & 3$\pm$2 \\

    CelebA & $44\pm7$ & $18\pm5$ & $4\pm2$ & $33\pm6$ & $12\pm3$ & $3\pm1$ & $52\pm21$ & $19\pm12$ & $4\pm3$ & $43\pm18$ & $14\pm1$ & $3\pm2$ \\

    CIFAR{-}10 &73$\pm$18 & 33$\pm$19 & 5$\pm$5 & 62$\pm$8 & 24$\pm$15 & 4$\pm$2 & 50$\pm$21 & 13$\pm$10 & 1$\pm$2 & 42$\pm$16 & 10$\pm$6 & 1$\pm$1 \\
    \bottomrule
  \end{tabular}
  \end{adjustbox}
  \vspace*{-0.10in}
\end{table}
\textbf{The impact of the routing algorithms on interpretability: } In classification tasks for vision problems, which are central to our experiments, the routing algorithm used in the CapsNet structure plays a critical role in interpretability. Specifically, it influences how the model assigns relevant units to the parts and wholes for each class, enhancing the model’s ability to model spatial hierarchies.
Given a well-trained CapsNet, as in \cite{SF17}, our experiments indicate that routing coefficients can be leveraged to identify the most relevant units based on forward and backward path estimations, enabling effective visualizations (see section \ref{sec:result-LayerWiseOcclusion}).
As can be seen in Fig.~\ref{fig:topkPerfCelebCifar} and Fig.~\ref{fig:classfPerEMForPass2}, the classification performance varies with the number of selected relevant units $k$ across the layers by utilizing different routing mechanisms \cite{SF17,HSF18}.
Notably, however, when switching from the DR~\cite{SF17} to the EM~\cite{HSF18} routing algorithm, the network required slightly more units to achieve comparable performance.

Additionally, we analyzed the exclusive and shared units across different routing mechanisms to reveal differences in interpretability. Fig.~\ref{fig:sharedUnit} illustrates that each routing mechanism captures distinct characteristics of unique and shared units across the CapsNet architectures.
These observations suggest that the routing mechanisms effectively identify class-specific features within both architectures.

For input reconstruction, Fig.~\ref{fig:relevantPathDR_EM} clearly shows that when using dynamic routing, the reconstructions tend to exhibit more disconnected elements. In contrast, EM routing produces blurred but fully connected reconstructions.
We examine the effect of input reconstructions using recent routing mechanisms \cite{MSC21, liu2024capsule} as shown in Fig.\ref{fig:perWRTRelatedMethods}. Notably, the quality of the input reconstructions varies significantly with different routing mechanisms. Additionally, we observe that training a deeper model may yield improved reconstructions \cite{liu2024capsule}. 
Regarding the visual heatmaps, Fig.\ref{fig:semanticParts} and \ref{fig:partWholeEM} demonstrate that both algorithms can detect different parts (\textit{PC}) of the input image to model the whole (\textit{CC}). However, there are notable differences in the sparsity of the heatmaps, indicating that each algorithm captures and prioritizes features in varying degrees of concentration, which can influence their interpretability and focus on relevant regions.

\section{Challenges and Future Work}
CapsNets face significant challenges when applied to large-scale datasets (i.e., ImageNet), primarily due to their computational complexity. The dynamic routing mechanism proposed by \cite{SF17} involves iterative calculations across capsule layers, making the process computationally expensive and particularly slow when handling large and complex datasets. 

Training CapsNet architectures on large-scale datasets requires substantial computational resources due to the increased complexity introduced by the routing mechanism. Moreover, as the input image size increases, the number of units in intermediate layers also rises.
This results in a significant number of capsule units between the \textit{PC} and \textit{CC} layers, complicating both optimization and scalability.
Therefore, CapsNets struggle to perform effectively on large-scale datasets. This inherent complexity also limits the network's ability to reliably capture complex \textit{part-whole} relationships, particularly as the number of classes increases, making it challenging to analyze CapsNets' overall behavior and interpretability.

In terms of interpretability, our methodology suggests that the limited performance of CapsNets on large-scale datasets restricts their applicability in such contexts. 
Effective interpretability requires sufficient performance, as this is essential for accurately analyzing the network's behavior. This limitation is evident in our experiments with the CIFAR-10 dataset, where performance is noticeably reduced compared to simpler datasets like MNIST and SVHN but it is still acceptable. This decrease may be attributed to the higher complexity of objects within CIFAR-10 images.

\textbf{Future Directions: }
First, we aim to evaluate our methodology using other routing algorithms \cite{gu2020improving, lin2022feature, LKS23} and several CapsNet backbones. This ensures generalizability across a broad range of enhanced capsule-based methods, thereby increasing their interpretability.

Second, we will consider employing Grad-CAM as an explanation method to explain \textit{CapsNet} predictions. Comparing the results from Grad-CAM with those obtained using our current method could provide additional insight into the observed \textit{RMA} performance discussed in Section 6.3. Additionally, this comparison may help determine whether the issue lies in the sparsity of the heatmaps produced by the explanation methods from \cite{SV14} or the \textit{RMA} protocol itself. 

Third, the balance between interpretability and model complexity could be explored further to provide a more comprehensive understanding of such models.

Finally, we encourage researchers to develop CapsNet architectures that reduce the overhead of the routing mechanism, as this could enable more effective analysis of their behavior on complex datasets, such as ImageNet and MS COCO. 
Additionally, deeper CapsNet architectures may be necessary for training larger datasets.

\section{Conclusion}
\label{sec:conclusions}
We propose a methodology to assess the interpretation properties of capsule networks.
Our analysis is centered on the identification and ablation of relevant units in the network. 
Our qualitative visualizations and quantitative results suggest that the representation encoded in CapsNets might not be as disentangled nor explicitly related to \textit{part-whole} features as is usually stated in the literature. 
Future work will concentrate in a denser analysis of the architecture plus pinpointing the effect that the selection of k-top units has in the obtained insights.
We hope the proposed methodology and discussed observations serve as a starting point for future efforts toward a deeper study and understanding of representations learned via CapsNets.

\section*{Declaration of Competing Interest}
The authors declare that they have no competing financial interests or personal relationships that could have appeared to influence the work reported in this paper. 

\section*{Code Availability \& Data}
Experiments related to this work were conducted on the publicly available MNIST, SVHN, CIFAR{-}10, and CelebA datasets~\cite{deng2012mnist,goodfellow2013multi, alex2009learning,LLW20}. The code is publicly available in the following link.\footnote{\url{https://github.com/STawalbeh/Representations-Learned-via-Capsule-based-Network-Architectures}}.

\bibliographystyle{elsarticle-num} 
\bibliography{casrefs}

\begin{thebibliography}{10}
\expandafter\ifx\csname url\endcsname\relax
  \def\url#1{\texttt{#1}}\fi
\expandafter\ifx\csname urlprefix\endcsname\relax\def\urlprefix{URL }\fi
\expandafter\ifx\csname href\endcsname\relax
  \def\href#1#2{#2} \def\path#1{#1}\fi

\bibitem{SF17}
S.~Sabour, N.~Frosst, G.~E. Hinton, Dynamic routing between capsules, Advances in neural information processing systems 30 (2017).

\bibitem{deepika2022improved}
J.~Deepika, C.~Rajan, T.~Senthil, Improved capsnet model with modified loss function for medical image classification, Signal, Image and Video Processing 16~(8) (2022) 2269--2277.

\bibitem{afriyie2022classification}
Y.~Afriyie, B.~A.~Weyori, A.~A.~Opoku, Classification of blood cells using optimized capsule networks, Neural Processing Letters 54~(6) (2022) 4809--4828.

\bibitem{WWLLT23}
L.~Wang, M.~Tang, X.~Hu, Evaluation of grouped capsule network for intracranial hemorrhage segmentation in ct scans, Scientific Reports 13~(1) (2023) 3440.

\bibitem{kim2020}
J.~Kim, S.~Jang, E.~Park, S.~Choi, Text classification using capsules, Neurocomputing 376 (2020) 214--221.

\bibitem{cheng2022hsan}
Y.~Cheng, H.~Zou, H.~Sun, H.~Chen, Y.~Cai, M.~Li, Q.~Du, Hsan-capsule: A novel text classification model, Neurocomputing 489 (2022) 521--533.

\bibitem{lin2022feature}
Z.~Lin, J.~Jia, F.~Huang, W.~Gao, Feature correlation-steered capsule network for object detection, Neural Networks 147 (2022) 25--41.

\bibitem{yu2021sparse}
Y.~Yu, J.~Wang, H.~Qiang, M.~Jiang, E.~Tang, C.~Yu, Y.~Zhang, J.~Li, Sparse anchoring guided high-resolution capsule network for geospatial object detection from remote sensing imagery, International Journal of Applied Earth Observation and Geoinformation 104 (2021) 102548.

\bibitem{liu2022disentangled}
Y.~Liu, D.~Zhang, N.~Liu, S.~Xu, J.~Han, Disentangled capsule routing for fast part-object relational saliency, IEEE Transactions on Image Processing 31 (2022) 6719--6732.

\bibitem{liu2024capsule}
Y.~Liu, D.~Cheng, D.~Zhang, S.~Xu, J.~Han, Capsule networks with residual pose routing, IEEE Transactions on Neural Networks and Learning Systems (2024).

\bibitem{liu2021part}
Y.~Liu, D.~Zhang, Q.~Zhang, J.~Han, Part-object relational visual saliency, IEEE transactions on pattern analysis and machine intelligence 44~(7) (2021) 3688--3704.

\bibitem{liu2024deep}
Y.~Liu, X.~Dong, D.~Zhang, S.~Xu, Deep unsupervised part-whole relational visual saliency, Neurocomputing 563 (2024) 126916.

\bibitem{Hyperspectral}
F.~Deng, S.~Pu, X.~Chen, Y.~Shi, T.~Yuan, S.~Pu, Hyperspectral image classification with capsule network using limited training samples, Sensors 18~(9) (2018) 3153.

\bibitem{wang2018hyperspectral}
W.-Y. Wang, H.-C. Li, L.~Pan, G.~Yang, Q.~Du, Hyperspectral image classification based on capsule network, in: IGARSS 2018-2018 IEEE International Geoscience and Remote Sensing Symposium, IEEE, 2018, pp. 3571--3574.

\bibitem{wang2022}
J.~Wang, X.~Tan, J.~Lai, J.~Li, Aspcnet: Deep adaptive spatial pattern capsule network for hyperspectral image classification, Neurocomputing 486 (2022) 47--60.

\bibitem{HI18}
J.~Heylen, S.~Iven, B.~De~Brabandere, J.~Oramas, L.~Van~Gool, T.~Tuytelaars, From pixels to actions: Learning to drive a car with deep neural networks, in: 2018 IEEE Winter Conference on Applications of Computer Vision (WACV), IEEE, 2018, pp. 606--615.

\bibitem{grigorescu2020survey}
S.~Grigorescu, B.~Trasnea, T.~Cocias, G.~Macesanu, A survey of deep learning techniques for autonomous driving, Journal of field robotics 37~(3) (2020) 362--386.

\bibitem{2020video}
D.~Liu, Y.~Cui, Y.~Chen, J.~Zhang, B.~Fan, Video object detection for autonomous driving: Motion-aid feature calibration, Neurocomputing 409 (2020) 1--11.

\bibitem{sezer2020financial}
O.~B. Sezer, M.~U. Gudelek, A.~M. Ozbayoglu, Financial time series forecasting with deep learning: A systematic literature review: 2005--2019, Applied soft computing 90 (2020) 106181.

\bibitem{mukhometzianov2018capsnet}
R.~Mukhometzianov, J.~Carrillo, Capsnet comparative performance evaluation for image classification, arXiv preprint arXiv:1805.11195 (2018).

\bibitem{JL20}
D.~Jung, J.~Lee, J.~Yi, S.~Yoon, icaps: An interpretable classifier via disentangled capsule networks, in: European Conference on Computer Vision, Springer, 2020, pp. 314--330.

\bibitem{FV17}
R.~C. Fong, A.~Vedaldi, Interpretable explanations of black boxes by meaningful perturbation, in: Proceedings of the IEEE international conference on computer vision, 2017, pp. 3429--3437.

\bibitem{GR16}
F.~Gr{\"u}n, C.~Rupprecht, N.~Navab, F.~Tombari, A taxonomy and library for visualizing learned features in convolutional neural networks, ICML, 2016.

\bibitem{HA16}
L.~A. Hendricks, Z.~Akata, M.~Rohrbach, J.~Donahue, B.~Schiele, T.~Darrell, Generating visual explanations, in: Computer Vision--ECCV 2016: 14th European Conference, Amsterdam, The Netherlands, October 11--14, 2016, Proceedings, Part IV 14, Springer, 2016, pp. 3--19.

\bibitem{ZF14}
M.~D. Zeiler, R.~Fergus, Visualizing and understanding convolutional networks, in: ECCV, 2014.

\bibitem{BK17}
D.~Bau, B.~Zhou, A.~Khosla, A.~Oliva, A.~Torralba, Network dissection: Quantifying interpretability of deep visual representations, in: Proceedings of the IEEE conference on computer vision and pattern recognition, 2017, pp. 6541--6549.

\bibitem{OW19}
J.~Oramas, K.~Wang, T.~Tuytelaars, Visual explanation by interpretation: Improving visual feedback capabilities of deep neural networks, in: ICLR, 2019.

\bibitem{SV14}
K.~Simonyan, A.~Vedaldi, A.~Zisserman, Deep inside convolutional networks: Visualising image classification models and saliency maps, ICLR (2013).

\bibitem{selvaraju2017grad}
R.~R. Selvaraju, M.~Cogswell, A.~Das, R.~Vedantam, D.~Parikh, D.~Batra, Grad-cam: Visual explanations from deep networks via gradient-based localization, in: Proceedings of the IEEE international conference on computer vision, 2017, pp. 618--626.

\bibitem{ZW18}
Q.~Zhang, Y.~N. Wu, S.-C. Zhu, Interpretable convolutional neural networks, in: Proceedings of the IEEE conference on computer vision and pattern recognition, 2018, pp. 8827--8836.

\bibitem{LQ19}
C.~Li, C.~Quan, L.~Peng, Y.~Qi, Y.~Deng, L.~Wu, A capsule network for recommendation and explaining what you like and dislike, in: Proceedings of the 42nd international ACM SIGIR conference on research and development in information retrieval, 2019, pp. 275--284.

\bibitem{WH20}
Z.~Wang, X.~Hu, S.~Ji, icapsnets: Towards interpretable capsule networks for text classification, CoRR (2020).

\bibitem{SG18}
Y.~Shen, M.~Gao, Dynamic routing on deep neural network for thoracic disease classification and sensitive area localization, in: Machine Learning in Medical Imaging: 9th International Workshop, MLMI 2018, Held in Conjunction with MICCAI 2018, Granada, Spain, September 16, 2018, Proceedings 9, Springer, 2018, pp. 389--397.

\bibitem{SH21}
Y.~Shi, L.~Han, W.~Huang, S.~Chang, Y.~Dong, D.~Dancey, L.~Han, A biologically interpretable two-stage deep neural network (bit-dnn) for vegetation recognition from hyperspectral imagery, IEEE Transactions on Geoscience and Remote Sensing 60 (2021) 1--20.

\bibitem{JC18}
D.~R. de~Jesus, J.~Cuevas, W.~Rivera, S.~Crivelli, Capsule networks for protein structure classification and prediction, arXiv preprint arXiv:1808.07475 (2018).

\bibitem{SM18}
A.~Shahroudnejad, P.~Afshar, K.~N. Plataniotis, A.~Mohammadi, Improved explainability of capsule networks: Relevance path by agreement, in: 2018 ieee global conference on signal and information processing (globalsip), IEEE, 2018, pp. 549--553.

\bibitem{BA20}
A.~Bhullar, Interpreting capsule networks for classification by routing path visualization, Ph.D. thesis, University of Guelph (2020).

\bibitem{MKGS23}
M.~Mitterreiter, M.~Koch, J.~Giesen, S.~Laue, Why capsule neural networks do not scale: Challenging the dynamic parse-tree assumption, in: Proceedings of the AAAI Conference on Artificial Intelligence, Vol.~37, 2023, pp. 9209--9216.

\bibitem{3DZYGG23}
Y.~Zhao, G.~Fang, Y.~Guo, L.~Guibas, F.~Tombari, T.~Birdal, 3dpointcaps++: Learning 3d representations with capsule networks, International Journal of Computer Vision 130~(9) (2022) 2321--2336.

\bibitem{KS19}
A.~Kosiorek, S.~Sabour, Y.~W. Teh, G.~E. Hinton, Stacked capsule autoencoders, Advances in neural information processing systems 32 (2019).

\bibitem{HSF18}
G.~E. Hinton, S.~Sabour, N.~Frosst, Matrix capsules with em routing, in: International conference on learning representations (ICLR), 2018.

\bibitem{amer2020path}
M.~Amer, T.~Maul, Path capsule networks, Neural Processing Letters (2020).

\bibitem{LinCY13ICLR2014}
M.~Lin, Q.~Chen, S.~Yan, Network in network, in: ICLR, 2014.

\bibitem{MSC21}
V.~Mazzia, F.~Salvetti, M.~Chiaberge, Efficient-capsnet: Capsule network with self-attention routing, Scientific reports 11~(1) (2021) 14634.

\bibitem{GT20}
J.~Gu, Interpretable graph capsule networks for object recognition, in: Proceedings of the AAAI Conference on Artificial Intelligence, Vol.~35, 2021, pp. 1469--1477.

\bibitem{RS22}
H.~Ren, J.~Su, H.~Lu, Evaluating generalization ability of convolutional neural networks and capsule networks for image classification via top-2 classification, arXiv preprint arXiv:1901.10112 (2019).

\bibitem{NT20}
X.~Ning, W.~Tian, W.~Li, Y.~Lu, S.~Nie, L.~Sun, Z.~Chen, Bdars\_capsnet: Bi-directional attention routing sausage capsule network, IEEE Access 8 (2020) 59059--59068.

\bibitem{sprj22}
S.~Pawan, J.~Rajan, Capsule networks for image classification: A review, Neurocomputing (2022).

\bibitem{patrick2019capsule}
M.~K. Patrick, A.~F. Adekoya, A.~A. Mighty, B.~Y. Edward, Capsule networks--a survey, Journal of King Saud University-computer and information sciences 34~(1) (2022) 1295--1310.

\bibitem{ZP15}
Z.~Liu, P.~Luo, X.~Wang, X.~Tang, Deep learning face attributes in the wild, in: ICCV, 2015.

\bibitem{LLW20}
C.-H. Lee, Z.~Liu, L.~Wu, P.~Luo, Maskgan: Towards diverse and interactive facial image manipulation, in: IEEE/CVPR, 2020.

\bibitem{kingma2017adam}
D.~P. Kingma, J.~Ba, Adam: a method for stochastic optimization (2014), arXiv preprint arXiv:1412.6980 15 (2017).

\bibitem{GM18}
A.~Gonzalez-Garcia, D.~Modolo, V.~Ferrari, Do semantic parts emerge in convolutional neural networks?, International Journal of Computer Vision (2018).

\bibitem{AO22}
L.~Arras, A.~Osman, W.~Samek, Clevr-xai: A benchmark dataset for the ground truth evaluation of neural network explanations, Information Fusion (2022).

\bibitem{VandersmissenExpEval}
B.~Vandersmissen, J.~Oramas, On the coherence of quantitative evaluation of visual explanations, arXiv:2302.10764 (2023).

\bibitem{gu2020improving}
J.~Gu, V.~Tresp, Improving the robustness of capsule networks to image affine transformations, in: Proceedings of the IEEE/CVF conference on computer vision and pattern recognition, 2020, pp. 7285--7293.

\bibitem{LKS23}
X.~Luo, X.~Kang, A.~Sch{\"o}nhuth, Predicting the prevalence of complex genetic diseases from individual genotype profiles using capsule networks, Nature Machine Intelligence 5~(2) (2023) 114--125.

\bibitem{deng2012mnist}
L.~Deng, The mnist database of handwritten digit images for machine learning research [best of the web], IEEE signal processing magazine 29~(6) (2012) 141--142.

\bibitem{goodfellow2013multi}
I.~J. Goodfellow, Y.~Bulatov, J.~Ibarz, S.~Arnoud, V.~Shet, Multi-digit number recognition from street view imagery using deep convolutional neural networks, arXiv preprint arXiv:1312.6082 (2013).

\bibitem{alex2009learning}
K.~Alex, Learning multiple layers of features from tiny images, https://www. cs. toronto. edu/kriz/learning-features-2009-TR. pdf (2009).

\end{thebibliography}
\end{document}